\documentclass[sigconf,authorversion,nonacm,screen]{acmart}

\AtBeginDocument{%
  \providecommand\BibTeX{{%
    \normalfont B\kern-0.5em{\scshape i\kern-0.25em b}\kern-0.8em\TeX}}}

\usepackage{algorithm}
\usepackage{algpseudocode}
\usepackage{textcomp}
\usepackage{xcolor}
\usepackage{graphicx}
\usepackage{epstopdf}

\usepackage{amsbsy}
\usepackage{amsthm}
\usepackage{longtable}
\usepackage{stackengine}
\usepackage{mathtools}
\usepackage{stmaryrd}
\usepackage{relsize}
\usepackage{multirow}
\usepackage{bm}
\usepackage[T1]{fontenc}
\usepackage{caption}

\usepackage{subcaption}
\usepackage{dsfont}
\usepackage{url}
\usepackage{color}

\usepackage{xfrac}
\usepackage{nicefrac}
\usepackage{setspace}

\allowdisplaybreaks[1]  

\usepackage{array}
\usepackage{booktabs,tabularx}
\usepackage{arydshln}
\usepackage{colortbl}
\usepackage{stfloats}

\def \H {\mathrm{H}} 
\def \I {\mathrm{I}} 
\def \D {\mathrm{D}} 

\def\markov{\hbox{$\--$}\kern-1.5pt\hbox{$\circ$}\kern-1.5pt\hbox{$\--$}}

\begin{document}
\title[Variational Leakage]{Variational Leakage:\space The Role of\\Information Complexity in Privacy Leakage}
\author{Amir Ahooye Atashin}
\authornote{Both authors contributed equally to this research.}
\email{amir.atashin@gmail.com}
\orcid{0000-0002-6788-4002}
\authornotemark[0]
\affiliation{%
  \institution{University of Geneva}
  \city{Geneva}
  \country{Switzerland}
}

\author{Behrooz Razeghi}
\authornote{Work done while at Imperial College London.}
\email{behrooz.razeghi@unige.ch}
\orcid{0000-0001-9568-4166}
\authornotemark[1]
\affiliation{%
  \institution{University of Geneva}
  \city{Geneva}
  \country{Switzerland}
}

\author{Deniz G\"{u}nd\"{u}z}
\email{d.gunduz@imperial.ac.uk}
\orcid{0000-0002-6788-4002}
\affiliation{%
  \institution{Imperial College London}
  \city{London}
  \country{United Kingdom}
}

\author{Slava Voloshynovskiy}
\email{svolos@unige.ch}
\orcid{0000-0002-7725-395X}
\affiliation{%
  \institution{University of Geneva}
  \city{Geneva}
  \country{Switzerland}
}


\begin{abstract}
We study the role of information complexity in privacy leakage about an attribute of an adversary's interest, which is not known \textit{a priori} to the system designer. Considering the supervised representation learning setup and using neural networks to parameterize the variational bounds of information quantities, we study the impact of the following factors on the amount of information leakage: information complexity regularizer weight, latent space dimension, the cardinalities of the known utility and unknown sensitive attribute sets, the correlation between utility and sensitive attributes, and a potential bias in a sensitive attribute of adversary's interest. We conduct extensive experiments on Colored-MNIST and CelebA datasets to evaluate the effect of information complexity on the amount of intrinsic leakage.

{\if@ACM@anonymous\else\makeatother{A repository of the proposed method implementation, Colored-MNIST dataset generator and the corresponding analysis is publicly available at:

\url{https://github.com/BehroozRazeghi/Variational-Leakage}}\fi}
\end{abstract}


\maketitle

\section{Introduction}
\makeatletter
Sensitive information sharing is a challenging problem in information systems. It is often handled by obfuscating the available information before sharing it with other parties. In \cite{makhdoumi2014information}, this problem has been formalized as the \textbf{privacy funnel (PF)} in an information theoretic framework. Given two correlated random variables $\mathbf{S}$ and $\mathbf{X}$ with a joint distribution $P_{\mathbf{S}, \mathbf{X}}\,$, where $\mathbf{X}$ represents the available information and $\mathbf{S}$ the private latent variable, the goal of the PF model  is to find a representation $\mathbf{Z}$ of $\mathbf{X}$ using a stochastic mapping $P_{\mathbf{Z}\mid \mathbf{X}}$ such that: (i) $\mathbf{S} \markov \mathbf{X} \markov \mathbf{Z}$ form a Markov chain; and (ii) representation $\mathbf{Z}$ is maximally informative about the useful data $\mathbf{X}$ (maximizing Shannon's mutual information (MI) $\I \left( \mathbf{X}; \mathbf{Z} \right)$) while being minimally informative about the sensitive data $\mathbf{S}$ (minimizing $\I \left( \mathbf{S}; \mathbf{Z} \right)$). There have been many extensions of this model in the recent literature, e.g., \cite{makhdoumi2014information, calmon2015fundamental, basciftci2016privacy, sreekumar2019optimal, hsu2019obfuscation, rassouli2019data, rassouli2019optimal, razeghi2020perfectobfuscation, rassouli2020perfect_JSAIT}.
\makeatletter

%
\begin{figure}[!t]
\centering
\advance\leftskip-0.7cm
\advance\rightskip-0.8cm
\includegraphics[scale=0.71]{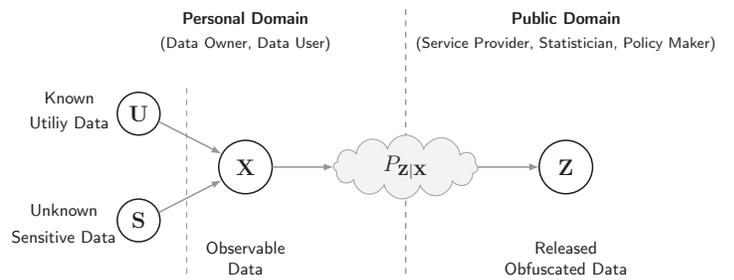}
\caption{The general setup.}
\label{Fig:First_diagram}
\end{figure}
%

In this paper, we will consider a delicate generalization of the PF model considered in \cite{basciftci2016privacy, rassouli2020perfect_JSAIT}, where the goal of the system designer is not to reveal the data that has available but another correlated utility variable. In particular, we assume that the data owner/user acquires some utility from the service provider based on the amount of information disclosed about a utility random variable $\mathbf{U}$ correlated with $\mathbf{X}$, measured by $\I \! \left( \mathbf{U}; \mathbf{Z} \right)$. Therefore, considering Markov chain $\left( \mathbf{U}, \mathbf{S}\right) \markov \mathbf{X} \markov \mathbf{Z}$,  the data owner's aim is to share a representation $\mathbf{Z}$ of \textit{observed data} $\mathbf{X}$, through a stochastic mapping $P_{\mathbf{Z} \mid \mathbf{X}}$, while preserving information about \textit{utility attribute} $\mathbf{U}$ and obfuscate information about \textit{sensitive attribute} $\mathbf{S}$ (see Fig.~\ref{Fig:First_diagram}).

The implicit assumption in the PF model presented above and the related generative adversarial privacy framework \cite{huang2017context, tripathy2019privacy} is to have \textit{pre-defined interests} in the game between the `defender' (data owner/user) and the `adversary'; that is, the data owner knows in advance what feature/ variable of the underlying data the adversary is interested in. Accordingly, the data release mechanism can be optimized/ tuned to minimize any inference the adversary can make about this specific random variable. However, this assumption is violated in most real-world scenarios. The attribute that the defender may assume as sensitive may not be the attribute of interest for the inferential adversary. As an example, for a given utility task at hand, the defender may try to restrict inference on gender recognition while the adversary is interested in inferring an individual's identity or facial emotion. Inspired by \cite{issa2019operational}, and in contrast to the above setups, we consider the scenario in which the adversary is curious about an attribute that is \textit{unknown} to the system designer.%

%
In particular, we argue that the information complexity of the representation measured by MI  $\I \! \left( \mathbf{X}; \mathbf{Z} \right)$ can also limit the information leakage about the unknown sensitive variable. In this paper, obtaining the parameterized variational approximation of information quantities, we investigate the core idea of \cite{issa2019operational} in the supervised representation learning setup.
\textbf{Notation:}
Throughout this paper, 
random vectors are denoted by capital bold letters (e.g., $\mathbf{X}$), deterministic vectors are denoted by small bold letters (e.g., $\mathbf{x}$), and alphabets (sets) are denoted by calligraphic fonts (e.g., $\mathcal{X}$). 
We use the shorthand $\left[ N \right]$ to denote the set $\{ 1, 2, \dots , N\}$. 
%
$\H \left( P_{\mathbf{X}} \right) \! \coloneqq \mathbb{E}_{P_{\mathbf{X}}} \left[ - \log P_{\mathbf{X}} \right]$ denotes the Shannon's entropy, while
$\H \left( P_{\mathbf{X}} \Vert Q_{\mathbf{X}} \right) \coloneqq \mathbb{E}_{P_{\mathbf{X}}} \left[ - \log Q_{\mathbf{X}} \right]$ denotes the cross-entropy of the distribution $P_{\mathbf{X}}$ relative to a distribution $Q_{\mathbf{X}}$.
\space The relative entropy is defined as $\D_{\mathrm{KL}}\left(P_{\mathbf{X}} \Vert Q_{\mathbf{X}}\right) \coloneqq \mathbb{E}_{P_{\mathbf{X}}} \big[ \log \frac{P_{\mathbf{X}}}{Q_{\mathbf{X}}} \big]$.%
\space The conditional relative entropy is defined by:
\begin{equation*}
\D_{\mathrm{KL}}\left( P_{\mathbf{Z}\mid \mathbf{X}}\Vert Q_{\mathbf{Z} \mid \mathbf{X}} \mid P_{\mathbf{X}}\right) \coloneqq \mathbb{E}_{P_{\mathbf{X}}} \left[\D_{\mathrm{KL}}\left( P_{\mathbf{Z}\mid\mathbf{X}=\mathbf{x}}\Vert Q_{\mathbf{Z}\mid\mathbf{X}=\mathbf{x}}\right)\right].
\end{equation*}
\noindent And the MI is defined by:
\begin{equation*}
\I \left( P_{\mathbf{X}}; P_{\mathbf{Z} \mid \mathbf{X}} \right)\coloneqq \D_{\mathrm{KL}} \left( P_{\mathbf{Z} \mid \mathbf{X}} \Vert P_{\mathbf{Z}}\mid P_{\mathbf{X}}\right)
\end{equation*}

\noindent We abuse notation to write $\H \left( \mathbf{X} \right) = \H \left( P_{\mathbf{X}} \right)$ and $\I \left( \mathbf{X}; \mathbf{Z} \right) = \I \left( P_{\mathbf{X}}; P_{\mathbf{Z} \mid \mathbf{X}} \right)$
for random vectors $\mathbf{X} \sim P_{\mathbf{X}}$ and $\mathbf{Z} \sim P_{\mathbf{Z}}$.
%

%
%
%
\section{Problem Formulation}\label{Sec:ProblemFormulation}

Given the observed data $\mathbf{X}$, the data owner wishes to release a representation $\mathbf{Z}$ for a utility task $\mathbf{U}$. Our aim is to investigate the potential statistical inference about a sensitive random attribute $\mathbf{S}$ from the released representation $\mathbf{Z}$. The sensitive attribute $\mathbf{S}$ is possibly also correlated with $\mathbf{U}$ and $\mathbf{X}$.

The objective is to obtain a stochastic map $P_{\mathbf{Z} \mid \mathbf{X}}: \! \mathcal{X} \rightarrow \mathcal{Z}$ such that $P_{\mathbf{U} \mid \mathbf{Z}} \! \approx \! P_{\mathbf{U} \mid \mathbf{X}}, \forall \, \mathbf{Z} \! \in \! \mathcal{Z}, \forall \, \mathbf{U} \! \in \! \mathcal{U}, \forall \, \mathbf{X} \! \in \! \mathcal{X}$.
This means that the posterior distribution of the utility attribute $\mathbf{U}$ is similar when conditioned on the released representation $\mathbf{Z}$ or on the original data $\mathbf{X}$. 
Under logarithmic loss, one can measure the utility by Shannon's MI \cite{makhdoumi2014information, tishby2000information, razeghi2020perfectobfuscation}. The logarithmic loss function has been widely used in learning theory \cite{cesa2006prediction}, image processing \cite{andre2006entropy}, information bottleneck \cite{harremoes2007information}, multi-terminal source coding \cite{courtade2011multiterminal}, and PF \cite{makhdoumi2014information}.

\textbf{Threat Model:} 
We make minimal assumptions about the adversary's goal, which can model a large family of potential adversaries. 
In particular, we have the following assumptions:
\begin{itemize}
\item
The distribution $P_{\mathbf{S} \mid \mathbf{X}}$ is unknown to the data user/owner. We only restrict attribute $\mathbf{S}$ to be discrete, which captures most scenarios of interest, e.g., a facial attribute, an identity, a political preference. 
\item
The adversary observes released representation $\mathbf{Z}$ and the Markov chain $(\mathbf{U}, \mathbf{S}) \markov \mathbf{X} \markov \mathbf{Z}$ holds. 
\item
We assume the adversary knows the mapping $P_{\mathbf{Z} \mid \mathbf{X}}$ designed by the data owner, i.e., the data release mechanism is public. Furthermore, the adversary may have access to a collection of the original dataset with the corresponding labels $\mathbf{S}$. 
\end{itemize}

Suppose that the sensitive attribute $\mathbf{S} \! \in \! \mathcal{S}$ has a uniform distribution over a discrete set $\mathcal{S}$, where $\vert \mathcal{S} \vert \! = \! 2^{L} \! < \! \infty$. If $\I (\mathbf{S}; \mathbf{Z}) \geq L - \epsilon$, then equivalently $\H (\mathbf{S}\!  \mid \! \mathbf{Z} ) \leq \epsilon$. Also note that due to the Markov chain $\mathbf{S} \markov \mathbf{X} \markov \mathbf{Z}$, we have $\I (\mathbf{S}; \mathbf{Z}) = \I (\mathbf{X}; \mathbf{Z}) - \I (\mathbf{X}; \mathbf{Z} \mid \mathbf{S})$. When $\mathbf{S}$ is not known a priori, the data owner has no control over $\I (\mathbf{X}; \mathbf{Z} \! \mid \! \mathbf{S})$. On the other hand, $\I (\mathbf{X}; \mathbf{Z})$ can be interpreted as the information complexity of the released representation, which plays a critical role in controlling the information leakage $\I (\mathbf{S}; \mathbf{Z})$. 
Note also that a statistic $\mathbf{Z} \! = \! f \!\left( \mathbf{X}\right)$ induces a partition on the sample space $\mathcal{X}$, where $\mathbf{Z}$ is sufficient statistic for $\mathbf{U}$ if and only if the assigned samples in each partition do not depend on $\mathbf{U}$. 
Hence, intuitively, a larger $\vert \mathcal{U} \vert$ induces finer partitions on $\mathcal{X}$, which could potentially lead to more leakage about the unknown random function $\mathbf{S}$ of $\mathbf{X}$. This is the core concept of the notion of \textit{variational leakage}, which we shortly address in our experiments.


Since the data owner does not know the particular sensitive variable of interest to the adversary, we argue that it instead aims to design $P_{\mathbf{Z} \mid \mathbf{X}}$ with the minimum (information) complexity and minimum utility loss. 
With the introduction of a Lagrange multiplier $\beta \! \in \! \left[ 0, 1 \right]$, we can formulate the objective of the data owner by \textit{maximizing} the associated~Lagrangian~functional:
\begin{equation}
    \mathcal{L} \left( P_{\mathbf{Z} \mid \mathbf{X}}, \beta   \right) = 
    \I  \left( \mathbf{U}; \mathbf{Z} \right) - \,  \beta  \, \I \left( \mathbf{X}; \mathbf{Z} \right). 
\end{equation}

This is the well-known \textbf{information bottleneck (IB)} principle \cite{tishby2000information}, which formulates the problem of extracting, in the most succinct way, the relevant information from random variable $\mathbf{X}$ about the random variable of interest $\mathbf{U}$. Given two correlated random variables $\mathbf{U}$ and $\mathbf{X}$ with joint distribution $P_{\mathbf{U,X}}$, the goal is to find a representation $\mathbf{Z}$ of $\mathbf{X}$ using a stochastic mapping $P_{\mathbf{Z}\mid \mathbf{X}}$ such that: (i) $\mathbf{U} \markov \mathbf{X} \markov \mathbf{Z}$, and (ii) $\mathbf{Z}$ is maximally informative about $\mathbf{U}$ (maximizing $\I \left( \mathbf{U}; \mathbf{Z} \right)$) and minimally informative about $\mathbf{X}$ (minimizing $\I \left( \mathbf{X}; \mathbf{Z}\right)$).



Note that in the PF model, $\I \left( \mathbf{X}; \mathbf{Z} \right)$ measures the \textit{useful} information, which is of the designer's interest, while in the IB model, $\I \left( \mathbf{U}; \mathbf{Z} \right)$ measures the \textit{useful} information. Hence, $\I \left( \mathbf{X}; \mathbf{Z} \mid \mathbf{S} \right)$ in PF quantifies the \textit{residual} information, while $\I \left( \mathbf{X}; \mathbf{Z} \mid \mathbf{U} \right)$ in IB quantifies the \textit{redundant} information.


In the sequel, we provide the parameterized variational approximation of information quantities, and then study the impact of the information complexity $\I \left( \mathbf{X}; \mathbf{Z} \right)$ on the information leakage for an unknown sensitive variable.

\subsection{Variational Approximation of Information Measures}


Let $Q_{\mathbf{U}\mid \mathbf{Z}} \! : \! \mathcal{Z} \! \rightarrow \! \mathcal{P}\!\left( \mathcal{U}\right)$, $Q_{\mathbf{S}\mid \mathbf{Z}} \! : \! \mathcal{Z}  \! \rightarrow \! \mathcal{P}\!\left( \mathcal{S}\right)$,  $Q_{\mathbf{Z}} \! : \! \mathcal{Z} \! \rightarrow \! \mathcal{P}\! \left( \mathcal{Z}\right)$ be variational approximations of the optimal utility decoder distribution $P_{\mathbf{U} \mid \mathbf{Z}}$, adversary decoder distribution $P_{\mathbf{S} \mid \mathbf{Z}}$, and latent space distribution $P_{\mathbf{Z}}$, respectively. 
The common approach is to use \textbf{deep neural networks (DNNs)} to model/parameterized these distributions. 
Let $P_{\boldsymbol{\phi}} (\mathbf{Z} \!\! \mid \!\! \mathbf{X})$ denote the family of encoding probability distributions $P_{\mathbf{Z} \mid \mathbf{X}}$ over $\mathcal{Z}$ for each element of space $\mathcal{X}$, parameterized by the output of a DNN $f_{\boldsymbol{\phi}}$ with parameters $\boldsymbol{\phi}$. 
Analogously, let $P_{\boldsymbol{\theta}} (\mathbf{U} \! \! \mid \! \! \mathbf{Z})$ and $P_{\boldsymbol{\xi}} \! \left( \mathbf{S} \! \mid \! \mathbf{Z} \right)$ denote the corresponding family of decoding probability distributions $Q_{\mathbf{U} \mid \mathbf{Z}}$ and $Q_{\mathbf{S} \mid \mathbf{Z}}$, respectively, parameterized by the output of DNNs $g_{\boldsymbol{\theta}}$ and $g_{\boldsymbol{\xi}}$. 
Let $P_{\mathsf{D}} \! \left( \mathbf{X} \right) \! = \! \frac{1}{N} \sum_{n=1}^{N} \! \delta ( \mathbf{x} - \mathbf{x}_n )$, $\mathbf{x}_n \in \mathcal{X}$ denote the empirical data distribution.
In this case, $P_{\boldsymbol{\phi}} \! \left( \mathbf{X}, \mathbf{Z} \right) \! = \! P_{\mathsf{D}} (\mathbf{X}) P_{\boldsymbol{\phi}} \! \left( \mathbf{Z} \! \mid \!  \mathbf{X} \right)$ denotes our joint inference data distribution, and $P_{\boldsymbol{\phi}} (\mathbf{Z}) \! = \! \mathbb{E}_{P_{\mathsf{D}} (\mathbf{X})} \left[ P_{\boldsymbol{\phi}} (\mathbf{Z} \! \mid \! \mathbf{X}) \right]$ denotes the learned \textit{aggregated} posterior distribution over latent space $\mathcal{Z}$.


\noindent
\textbf{Information Complexity:}
The information complexity can be decomposed as:
\begin{eqnarray}\label{I_xz_decomposition}
\!\! \I \left( \mathbf{X}; \mathbf{Z} \right)  \!\!\! \! &=&  \!\!\! \!  \mathbb{E}_{P_{\mathbf{X}, \mathbf{Z}}} \Big[ \log \frac{P_{\mathbf{X}, \mathbf{Z} }}{P_{\mathbf{X}} P_{\mathbf{Z}}} \Big]   \! = \!    
\mathbb{E}_{P_{\mathbf{X}, \mathbf{Z}}} \Big[ \log \frac{P_{\mathbf{Z} \mid \mathbf{X} }}{ Q_{\mathbf{Z}}}  \frac{Q_{\mathbf{Z}}}{P_{\mathbf{Z}}} \Big] \nonumber \\
&=&  \!\!\! \mathbb{E}_{P_{\mathbf{X}}} \left[ \D_{\mathrm{KL}}  \left( P_{\mathbf{Z} \mid \mathbf{X}} \Vert Q_{\mathbf{Z}} \right) \right] \! - \! \D_{\mathrm{KL}} \left( P_{\mathbf{Z}} \Vert Q_{\mathbf{Z}} \right). 
\end{eqnarray}
Where $Q_{\mathbf{Z}}$ is the latent space's prior.

Therefore, the parameterized variational approximation of information complexity \eqref{I_xz_decomposition} can be recast~as:
%
\begin{align}\label{Eq:I_XZ_phi}
\I_{\boldsymbol{\phi}} \! \left( \mathbf{X}; \mathbf{Z} \right) \coloneqq  \D_{\mathrm{KL}} \! \left( P_{\boldsymbol{\phi}} (\mathbf{Z} \! \mid \! \mathbf{X}) \, \Vert \, Q_{\mathbf{Z}} \mid P_{\mathsf{D}}(\mathbf{X}) \right) -  \D_{\mathrm{KL}} \! \left( P_{\boldsymbol{\phi}}(\mathbf{Z}) \, \Vert \, Q_{\mathbf{Z}} \right).  
\end{align}
%
The optimal prior $Q^{\ast}_{\mathbf{Z}}$ minimizing the information complexity is $Q^{\ast}_{\mathbf{Z}} (\mathbf{z}) = \mathbb{E}_{P_{\mathsf{D}} (\mathbf{X})} \left[ P_{\boldsymbol{\phi}} \left( \mathbf{Z} \mid \mathbf{X} = \mathbf{x} \right) \right]$; however, it may potentially lead to over-fitting.
A critical challenge is to guarantee that the learned aggregated posterior distribution $P_{\boldsymbol{\phi}} (\mathbf{Z})$ conforms well to thd prior $Q_{\mathbf{Z}}$ \cite{kingma2016improved, rezende2015variational, rosca2018distribution, tomczak2018vae, bauer2019resampled}. We can cope with this issue by employing a more \textit{expressive} form for $Q_{\mathbf{Z}}$,  which would allow us to provide a good fit of an arbitrary space for $\mathcal{Z}$, at the expense of additional \textit{computational complexity}.


\noindent
\textbf{Information Utility:}
The parameterized variational approximation of MI between the released representation $\mathbf{Z}$ and the utility attribute $\mathbf{U}$ can be recast~as:
\begin{equation}\label{Eq:I_UZ_phi_theta_SecondDecomposition}
\begin{split}
\I_{\boldsymbol{\phi}, \boldsymbol{\theta}}&\! \left( \mathbf{U}; \mathbf{Z} \right)\!\;\coloneqq \nonumber\\ 
& 
\mathbb{E}_{P_{\mathbf{U}, \mathbf{X}}} \Big[ \mathbb{E}_{P_{\boldsymbol{\phi}} \left( \mathbf{Z} \mid \mathbf{X} \right) } \Big[ \log \frac{P_{\boldsymbol{\theta}} \! \left( \mathbf{U} \! \mid \! \mathbf{Z} \right) }{P_{\mathbf{U}}}  \cdot \frac{P_{\boldsymbol{\theta}} (\mathbf{U})}{P_{\boldsymbol{\theta}} (\mathbf{U})} \Big] \Big]=\nonumber \\
&  
\mathbb{E}_{P_{\mathbf{U}, \mathbf{X}}} \left[  \mathbb{E}_{P_{\boldsymbol{\phi}} \left( \mathbf{Z} \mid \mathbf{X} \right) } \left[ \log P_{\boldsymbol{\theta}} \! \left( \mathbf{U} \! \mid \! \mathbf{Z} \right) \right] \right] \nonumber \\
& \qquad - \mathbb{E}_{P_{\mathbf{U}}} \big[ \log \frac{P_{\mathbf{U}}}{P_{\boldsymbol{\theta}} (\mathbf{U})}\big]
+ \mathbb{E}_{P_{\mathbf{U}}} \left[ \log P_{\boldsymbol{\theta}} (\mathbf{U}) \right] 
 \nonumber \\
& =    
- \H_{\boldsymbol{\phi}, \boldsymbol{\theta}} \left( \mathbf{U} \! \mid  \! \mathbf{Z} \right) 
- \D_{\mathrm{KL}} \left( P_{\mathbf{U}}  \Vert P_{\boldsymbol{\theta}} (\mathbf{U}) \right)
+ \H \left( P_{\mathbf{U}} \Vert P_{\boldsymbol{\theta}} (\mathbf{U}) \right) 
  \nonumber \\
& \geq  
 \underbrace{- \H_{\boldsymbol{\phi}, \boldsymbol{\theta}} \left( \mathbf{U} \! \mid  \! \mathbf{Z} \right) }_{\mathrm{Prediction~Fidelity}}
- \underbrace{ \D_{\mathrm{KL}} \left( P_{\mathbf{U}} \, \Vert \, P_{\boldsymbol{\theta}} (\mathbf{U}) \right)}_{\mathrm{Distribution~Discrepancy~Loss}} ,
\end{split}
\end{equation}
where $\H_{\boldsymbol{\phi}, \boldsymbol{\theta}} \!  \left( \mathbf{U} \! \mid  \! \mathbf{Z} \right)  \!   = \!  - \mathbb{E}_{P_{\mathbf{U}, \mathbf{X}}} \!  \left[  \mathbb{E}_{P_{\boldsymbol{\phi}} \left( \mathbf{Z} \mid \mathbf{X} \right) } \!  \left[ \log P_{\boldsymbol{\theta}} \! \left( \mathbf{U} \! \mid \! \mathbf{Z} \right) \right] \right]$ represents the parameterized decoder uncertainty, and in the last line we use the positivity of the cross-entropy $\H \left( P_{\mathbf{U}} \Vert P_{\boldsymbol{\theta}} (\mathbf{U}) \right)$.


%
%
%
\section{Learning Model}
\label{Sec:DeepVariationalApproximation}

\begin{figure}[!t]
\centering
\advance\leftskip-0.9cm
\advance\rightskip-0.9cm
\includegraphics[scale=0.56]{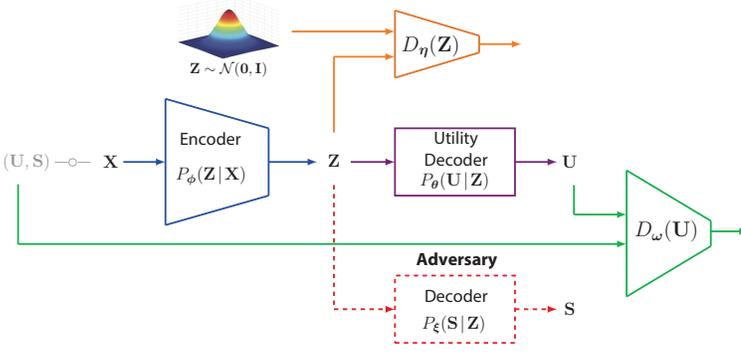}
\caption{The training and testing architecture. During training, the data user/owner trains the parameterized networks $\left( \boldsymbol{\phi}, \boldsymbol{\theta}, \boldsymbol{\eta}, \boldsymbol{\omega} \right)$. During testing, only the encoder-decoder pair $( \boldsymbol{\phi}, \boldsymbol{\theta} )$ is used. 
The adversary uses the publicly-known (fixed) encoder $\boldsymbol{\phi}$ and a collection of the original dataset, and trains an inference network $\boldsymbol{\xi}$ to infer attribute $\mathbf{S}$ of his interest.}
\label{Fig:Architecture}
\end{figure}

\textbf{System~Designer.} 
Given independent and identically distributed (i.i.d.) training samples $\{ \left( \mathbf{u}_n, \mathbf{x}_n  \right) \}_{n=1}^{N}$ $ \subseteq \mathcal{U} \times \mathcal{X}$, and using stochastic gradient descent (SGD)-type approach, DNNs $f_{\boldsymbol{\phi}}$, $g_{\boldsymbol{\theta}}$, $D_{\boldsymbol{\eta}}$, and $D_{\boldsymbol{\omega}}$ are trained together to maximize a Monte-Carlo approximation of the deep variational IB functional over parameters $\boldsymbol{\phi}$, $\boldsymbol{\theta}$, $\boldsymbol{\eta}$, and $\boldsymbol{\omega}$ (Fig.~\ref{Fig:Architecture}). Backpropagation through random samples from the posterior distribution $P_{\boldsymbol{\phi}} (\mathbf{Z} \! \!\mid \! \!\mathbf{X})$is required in our framework, which is a challenge since backpropagation cannot flow via random nodes; to overcome this hurdle, we apply the reparameterization approach \cite{kingma2014auto}).

The inferred posterior distribution is typically assumed to be a multi-variate Gaussian with a diagonal co-variance, i.e., $P_{\boldsymbol{\phi}} (\mathbf{Z} \! \mid \! \mathbf{x}) = \mathcal{N} \big( \boldsymbol{\mu}_{\boldsymbol{\phi}} (\mathbf{x}), $ $ \mathsf{diag} ( \boldsymbol{\sigma}_{\boldsymbol{\phi}} (\mathbf{x}) ) \big)$. 
Suppose $\mathcal{Z} = \mathbb{R}^d$. We first sample a random variable $\boldsymbol{\mathcal{E}}$ i.i.d. from $\mathcal{N} \! \left( \boldsymbol{0}, \mathbf{I}_d\right)$, then given data sample $\mathbf{x} \in \mathcal{X}$, we generate the sample $\mathbf{z} = \boldsymbol{\mu}_{\boldsymbol{\phi}} (\mathbf{x}) + \boldsymbol{\sigma}_{\boldsymbol{\phi}} (\mathbf{x}) \odot \boldsymbol{\varepsilon}$, where $\odot$ is the element-wise (Hadamard) product. 
%
%
The latent space prior distribution is typically considered as a fixed $d$-dimensional standard isotropic multi-variate Gaussian, i.e., $Q_{ \mathbf{Z}} = \mathcal{N} \! \left( \boldsymbol{0}, \mathbf{I}_d \right)$. 
For this simple choice, the information complexity upper bound 
\begin{equation*}
\mathbb{E}_{P_{\boldsymbol{\phi}} (\mathbf{X}, \mathbf{Z})} [ \log \frac{P_{\boldsymbol{\phi}} \left( \mathbf{Z}  \mid  \mathbf{X}\right) }{ Q_{ \mathbf{Z}}} ] = 
\mathbb{E}_{P_{\mathsf{D}} (\mathbf{X})} \left[ \D_{\mathrm{KL}}  \left( P_{\boldsymbol{\phi}} \! \left( \mathbf{Z} \! \mid \! \mathbf{X}\right) \Vert Q_{ \mathbf{Z}} \right) \right]
\end{equation*}

\noindent has a closed-form expression, which reads as:
\begin{equation*}
2 \, \D_{\mathrm{KL}} \left( P_{\boldsymbol{\phi}} \, \left( \mathbf{Z} \, \mid \, \mathbf{X}=\mathbf{x}\right)  \Vert \right. \left. Q_{ \mathbf{Z}} \right)={\Vert \boldsymbol{\mu}_{\boldsymbol{\phi}} (\mathbf{x}) \Vert}_2^2 + d + \sum_{i=1}^{d} ( \boldsymbol{\sigma}_{\boldsymbol{\phi}} (\mathbf{x})_i - \log \boldsymbol{\sigma}_{\boldsymbol{\phi}} (\mathbf{x})_i)
\end{equation*}

The $\mathrm{KL}$-divergences in \eqref{Eq:I_XZ_phi} and \eqref{Eq:I_UZ_phi_theta_SecondDecomposition} can be estimated using the density-ratio trick \cite{nguyen2010estimating, sugiyama2012density}, utilized in the GAN framework to directly match the data and generated model distributions. 
The trick is to express two distributions as conditional distributions, conditioned on a label $C \in \{ 0, 1 \}$, and reduce the task to binary classification. The key point is that we can estimate the KL-divergence by estimating the ratio of two distributions without modeling each distribution explicitly.

Consider $\D_{\mathrm{KL}} \! \left( P_{\boldsymbol{\phi}}(\mathbf{Z}) \, \Vert \, Q_{\mathbf{Z}} \right)= \mathbb{E}_{ P_{\boldsymbol{\phi}}(\mathbf{Z})}    [ \log \frac{ P_{\boldsymbol{\phi}}(\mathbf{Z})}{Q_{\mathbf{Z}}}  ]$. We now define $\rho_{\mathbf{Z}} (\mathbf{z}\! \mid \! c)$ as $\rho_{\mathbf{Z}} (\mathbf{z} \! \mid \! c\!=\!1) \! =\!  P_{\boldsymbol{\phi}}(\mathbf{Z})$, $\rho_{\mathbf{Z}} (\mathbf{z} \! \mid \! c\!=\!0) = Q_{\mathbf{Z}}$. 
Suppose that a perfect binary classifier (discriminator) $D_{\boldsymbol{\eta}} (\mathbf{z})$, with parameters $\boldsymbol{\eta}$, is trained to associate the label $c=1$ to samples from distribution $P_{\boldsymbol{\phi}}(\mathbf{Z}) $ and the label $c=0$ to samples from $Q_{\mathbf{Z}}$. Using the Bayes' rule and assuming that the marginal class probabilities are equal, i.e., $\rho (c=1) = \rho (c=0)$, the density ratio can be expressed as:
\begin{equation}\label{Eq:DensityRatioTrick}
    \frac{P_{\boldsymbol{\phi}}(\mathbf{Z} = \mathbf{z})}{Q_{\mathbf{Z}} ( \mathbf{z})} \! = \! 
    \frac{\rho_{\mathbf{Z}} (\mathbf{z} \mid c=1)}{\rho_{\mathbf{Z}} (\mathbf{z} \mid c=0)} \! = \!
    \frac{\rho_{\mathbf{Z}} (c=1 \mid \mathbf{z} )}{\rho_{\mathbf{Z}} ( c=0 \mid \mathbf{z})} \! \approx \! \frac{D_{\boldsymbol{\eta}} (\mathbf{z})}{1 - D_{\boldsymbol{\eta}} (\mathbf{z})}. \nonumber 
\end{equation}

Therefore, given a trained discriminator $D_{\boldsymbol{\eta}} (\mathbf{z})$ and $M$ i.i.d. samples $\{ \mathbf{z}_m \}_{m=1}^{M}$ from $P_{\boldsymbol{\phi}}(\mathbf{Z})$, we estimate $\D_{\mathrm{KL}} \! \left( P_{\boldsymbol{\phi}}(\mathbf{Z}) \, \Vert \, Q_{\mathbf{Z}}\right)$ as:
\begin{equation}\label{Eq:D_KL_Z_estimation}
    \D_{\mathrm{KL}} \! \left( P_{\boldsymbol{\phi}}(\mathbf{Z}) \, \Vert \, Q_{\mathbf{Z}} \right) \approx \frac{1}{M} \sum_{m=1}^{M} \log \frac{D_{\boldsymbol{\eta}} (\mathbf{z}_m)}{1 - D_{\boldsymbol{\eta}} (\mathbf{z}_m)}. 
\end{equation}
Our model is trained using alternating block coordinate descend across five steps (See Algorithm~\ref{Algorithm:VariationalNestedLeakage}).

\textbf{Inferential~Adversary:} 
Given the publicly-known encoder $\boldsymbol{\phi}$ and $K$ i.i.d. samples $\{ (\mathbf{s}_k , \mathbf{z}_k) \}_{k=1}^K \! \subseteq \mathcal{S} \times \mathcal{Z}$, the adversary trains an inference network $\boldsymbol{\xi}$ to minimize $\H_{\boldsymbol{\xi}} (\mathbf{S} \! \mid \! \mathbf{Z})$.

%
%
%
%
\begin{center}
\centering
\begin{spacing}{1}
\begin{algorithm}[!t]
    \setstretch{1.3}
    \begin{algorithmic}[1]
        \State \textbf{Inputs:} Training Dataset: $\{ \left( \mathbf{u}_n, \mathbf{x}_n  \right) \}_{n=1}^{N}$;
        
        \hspace{11pt}Hyper-Parameter: $\beta$;
        
        \State $\boldsymbol{\phi}, \boldsymbol{\theta}, \boldsymbol{\eta}, \boldsymbol{\omega}\; \gets$ Initialize Network Parameters
        
        \Repeat
        
        \hspace{-20pt}(1) {\small\textbf{\textsf{Train the Encoder and Utility Decoder}} $\left( \boldsymbol{\phi}, \boldsymbol{\theta} \right)$}
         \State  Sample a mini-batch $\{ \mathbf{u}_m, \mathbf{x}_m \}_{m=1}^{M} \sim P_{\mathsf{D}} (\mathbf{X}) P_{\mathbf{U} \mid \mathbf{X}}$
         \State  Compute $\mathbf{z}_m \sim f_{\boldsymbol{\phi}} (\mathbf{x}_m), \forall m \in [M]$
         \State  Back-propagate loss:\vspace{3pt}
         
            $ \mathcal{L} \! \left( \boldsymbol{\phi}, \! \boldsymbol{\theta} \right) \!  = \!  - \frac{1}{M} \! \sum_{m=1}^{M} \!\! \big( \log P_{\boldsymbol{\theta}} ( \mathbf{u}_m \! \mid  \! \mathbf{z}_m )$  
            
            $\qquad\qquad\qquad\quad  -  \beta \, \D_{\mathrm{KL}} \! \left( P_{\boldsymbol{\phi}} (\mathbf{z}_m \! \mid \! \mathbf{x}_m ) \Vert  Q_{\mathbf{Z}} (\mathbf{z}_m) \right)\! \big)$
                
        \vspace{3pt}  
        
        \hspace{-20pt}(2) {\small\textbf{\textsf{Train the Latent Space Discriminator}} $ \boldsymbol{\eta} $}
        \State  Sample $\{ \mathbf{x}_m \}_{m=1}^{M} \sim P_{\mathsf{D}} (\mathbf{X})$
        \State  Sample $\{ \mathbf{\widetilde{z}}_m \}_{m=1}^{M} \sim Q_{\mathbf{Z}}$
        \State  Compute $\mathbf{z}_m \sim f_{\boldsymbol{\phi}} (\mathbf{x}_m), \forall m \in [M]$
        \State  Back-propagate loss:\vspace{3pt}
        
                $\mathcal{L} \! \left( \boldsymbol{\eta} \right) \! \! = \! \! - \frac{ \beta}{M} \! \sum_{m=1}^{M} \!\! \big( \! \log \! D_{\boldsymbol{\eta}} (\mathbf{z}_m)  \! +\!  \log \! \left( 1 \! - \! D_{\boldsymbol{\eta}} ( \mathbf{\widetilde{z}}_m ) \right) )$
          
        \vspace{3pt}
        
        \hspace{-20pt}(3) {\small\textbf{\textsf{Train the Encoder $\boldsymbol{\phi}$ Adversarially}}}
        \State  Sample $\{ \mathbf{x}_m \}_{m=1}^{M} \sim P_{\mathsf{D}} (\mathbf{X})$
        \State  Compute $\mathbf{z}_m \sim f_{\boldsymbol{\phi}} (\mathbf{x}_m), \forall m \in [M]$
        \State  Back-propagate loss:  
                $\mathcal{L} \! \left( \boldsymbol{\phi} \right) \! = \! \!   \frac{ \beta}{M} \! \sum_{m=1}^{M} \! \log D_{\boldsymbol{\eta}} (\mathbf{z}_m)  \!$
          
        \vspace{3pt}
        
        \hspace{-20pt}(4) {\small\textbf{\textsf{Train the Attribute Class Discriminator}} $ \boldsymbol{\omega} $}
        \State  Sample $\{ \mathbf{u}_m \}_{m=1}^{M} \sim P_{\mathbf{U}} $
        \State  Sample $\{ \mathbf{\widetilde{z}}_m \}_{m=1}^{M} \sim Q_{\mathbf{Z}}$
        \State  Compute $\mathbf{\widetilde{u}}_m \sim g_{\boldsymbol{\theta}} \left( \mathbf{\widetilde{z}}_m \right), \forall m \in [M]$
        \State  Back-propagate loss:\vspace{3pt}
        
                \hspace{-4pt}$\mathcal{L} \! \left( \boldsymbol{\omega} \right) \! \! = \!\! - \frac{1}{M} \! \sum_{m=1}^{M}\! \big( \! \log \! D_{\boldsymbol{\omega}} (\mathbf{u}_m) \! + \!  \log \! \left( 1 \! - \! D_{\boldsymbol{\omega}} ( \mathbf{\widetilde{u}}_m ) \right))$
        
        \vspace{3pt}
        
        \hspace{-20pt}(5) {\small\textbf{\textsf{Train the Utility Decoder $\boldsymbol{\theta}$ Adversarially}}}
        \State  Sample $\{ \mathbf{\widetilde{z}}_m \}_{m=1}^{M} \sim Q_{\mathbf{Z}}$
        \State  Compute $\mathbf{\widetilde{u}}_m \sim g_{\boldsymbol{\theta}} \left( \mathbf{\widetilde{z}}_m \right), \forall m \in [M]$
        \State  Back-propagate loss:  
                $\mathcal{L} \! \left( \boldsymbol{\omega} \right) \! = \! \! \frac{1}{M} \sum_{m=1}^{M} \!  \log \left( 1 \! - \! D_{\boldsymbol{\omega}} ( \mathbf{\widetilde{u}}_m ) \right)$
        
        \vspace{3pt} 
        
        \Until{Convergence}
        \State \textbf{return} $\boldsymbol{\phi}, \boldsymbol{\theta}, \boldsymbol{\eta}, \boldsymbol{\omega}$
    \end{algorithmic}
    \caption{Training Algorithm: Data Owner}
    \label{Algorithm:VariationalNestedLeakage}
\end{algorithm}
\end{spacing}
\end{center}

%
%
%

\section{Experiments}

\begin{figure*}[!t]
\centering
\subfloat[\label{Fig:ColoredMNIST_a}]{%
  \includegraphics[width=0.30\textwidth]{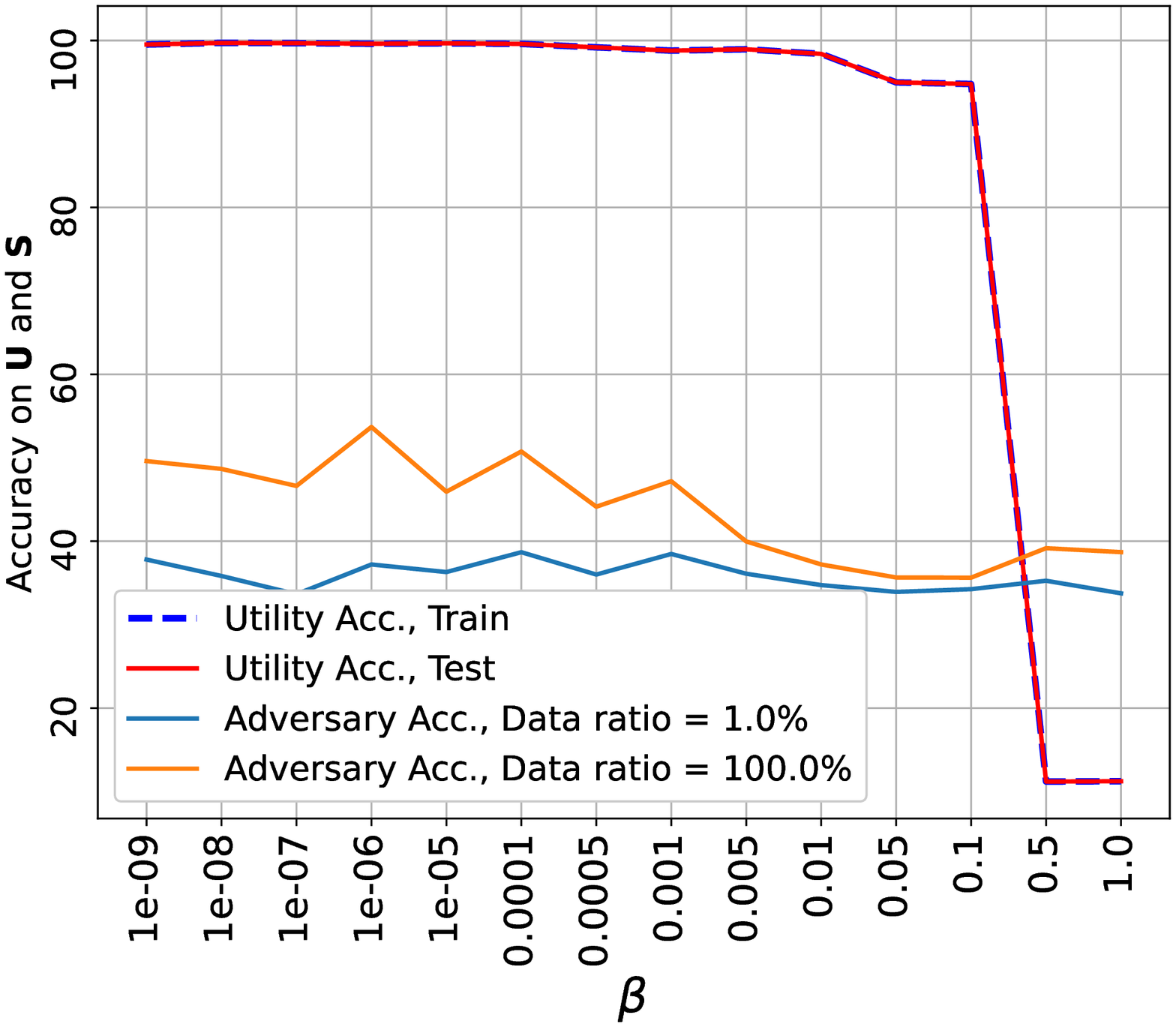}
}
\subfloat[\label{Fig:ColoredMNIST_b}]{%
  \includegraphics[width=0.30\textwidth]{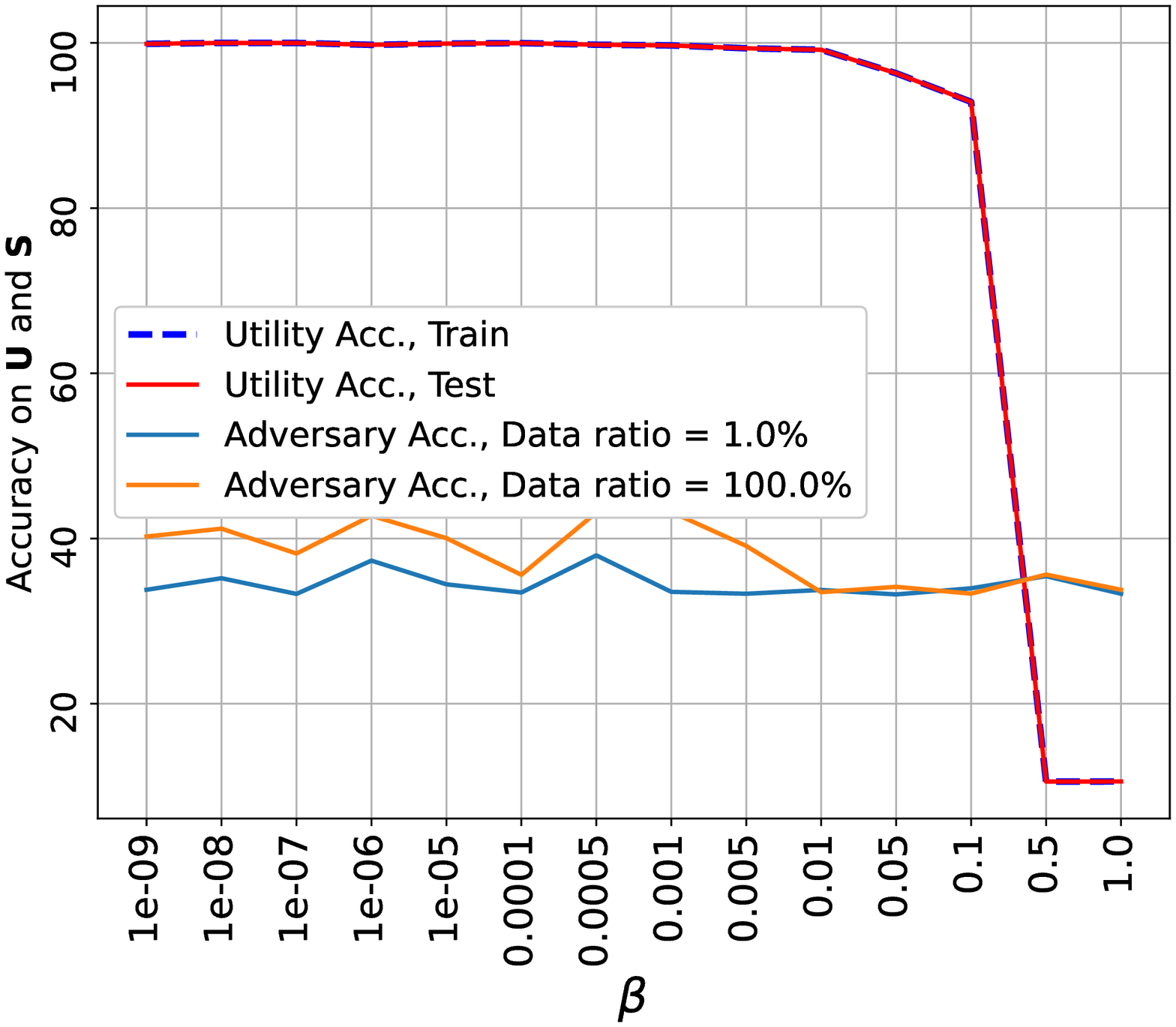}
}
\subfloat[\label{Fig:ColoredMNIST_c}]{%
  \includegraphics[width=0.30\textwidth]{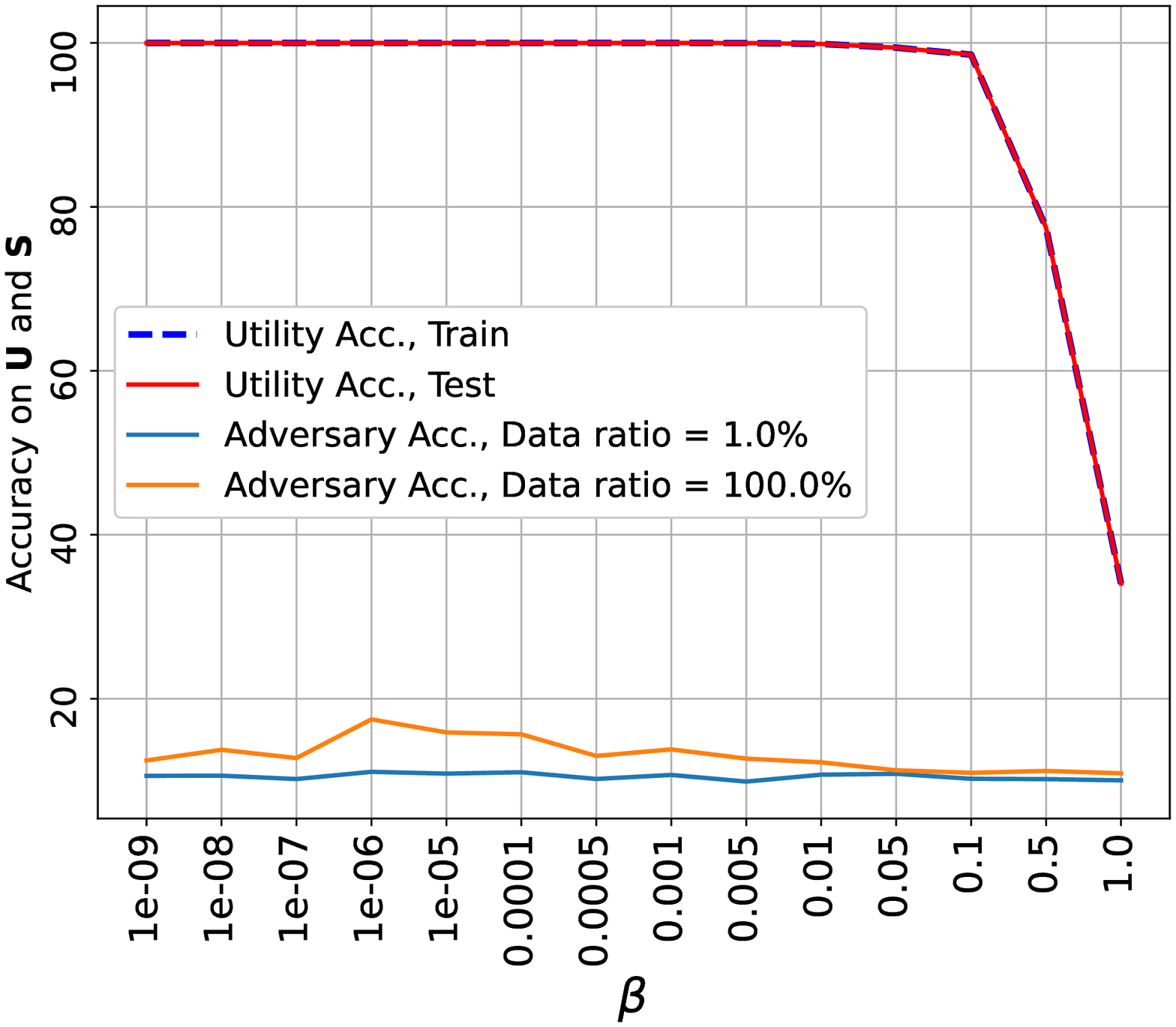}
}
\vfill{}

\subfloat[\label{Fig:ColoredMNIST_d}]{%
  \includegraphics[width=0.30\textwidth]{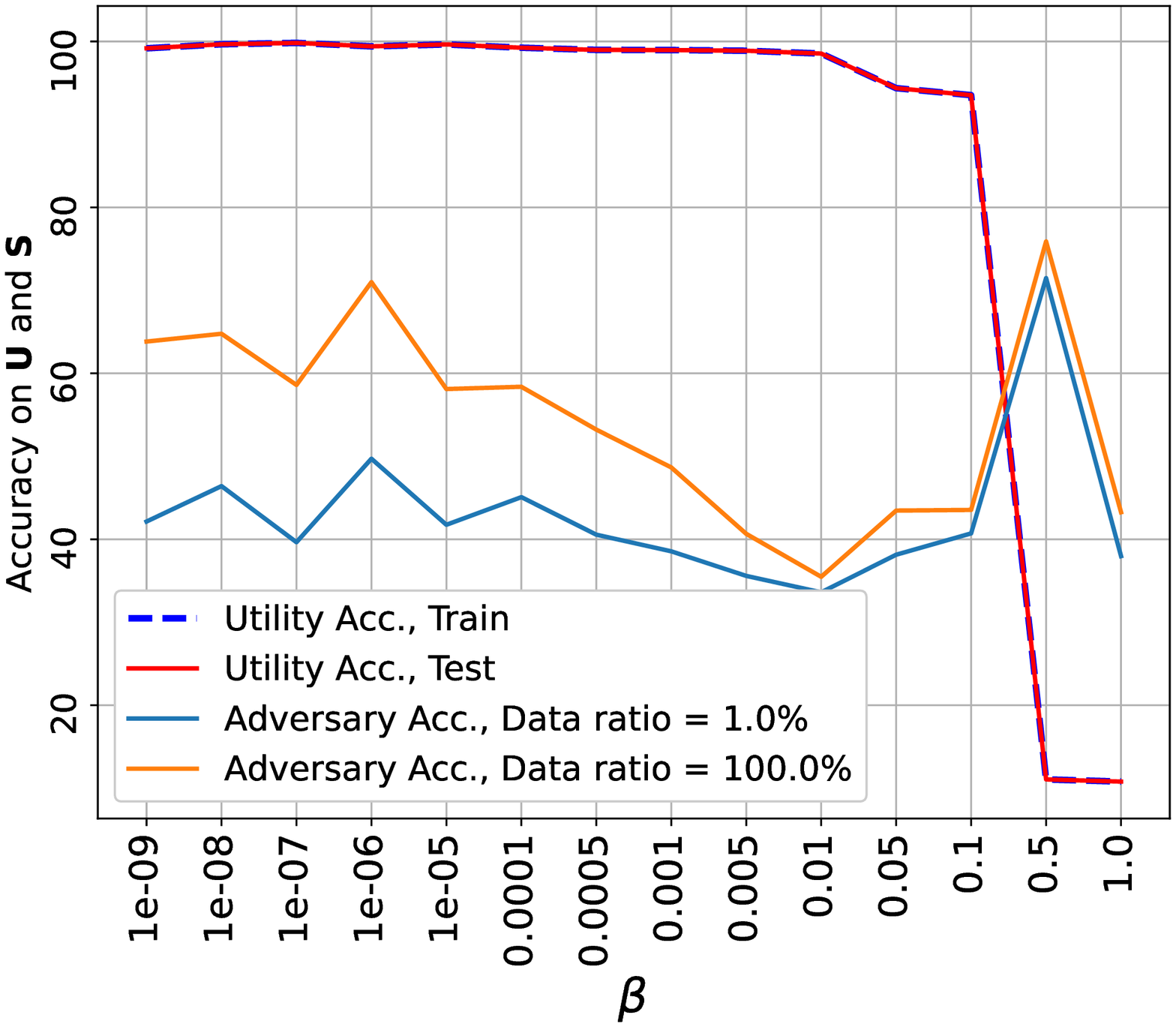}
}
\subfloat[\label{Fig:ColoredMNIST_e}]{%
  \includegraphics[width=0.30\textwidth]{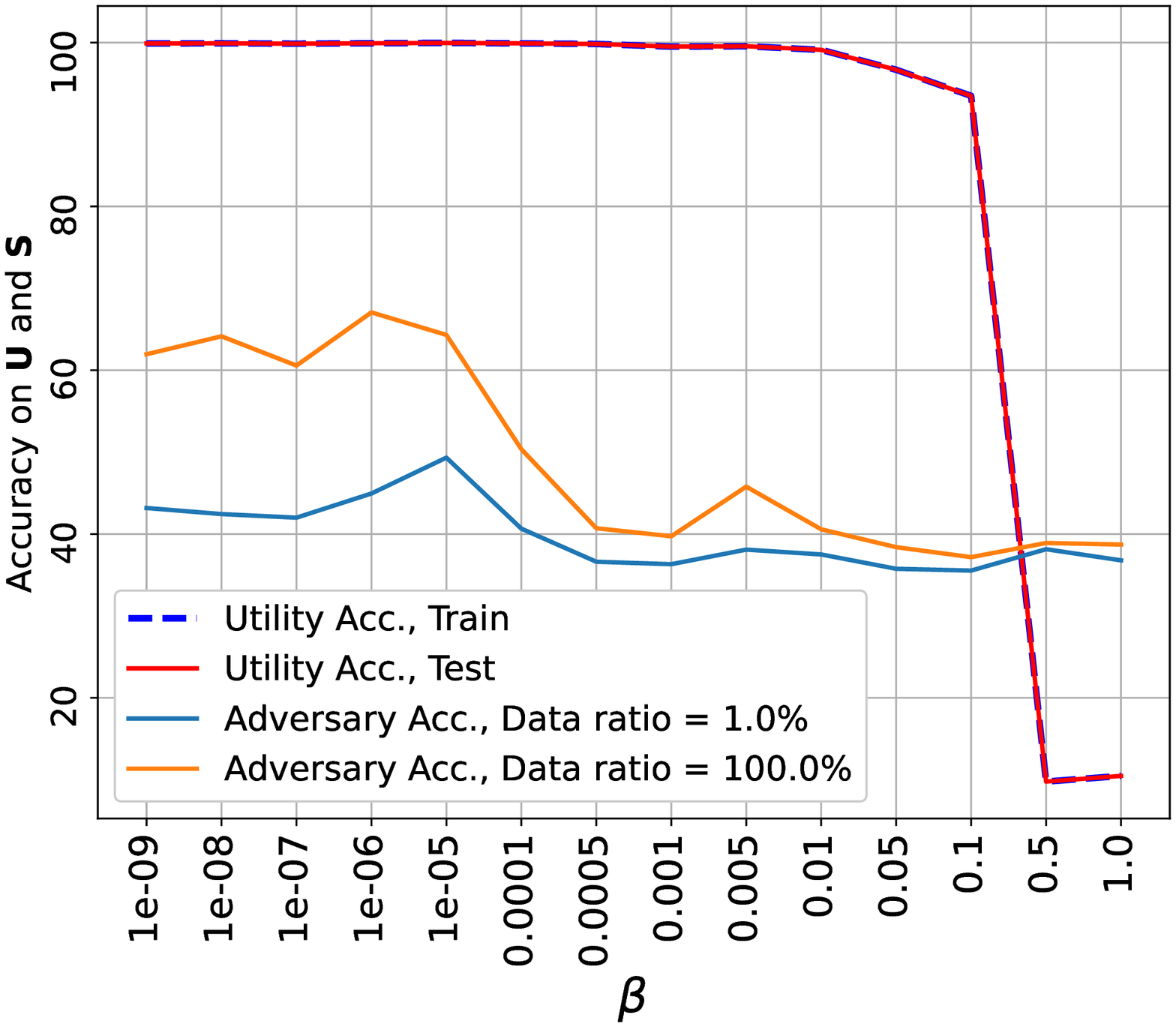}
}
\subfloat[\label{Fig:ColoredMNIST_f}]{%
  \includegraphics[width=0.30\textwidth]{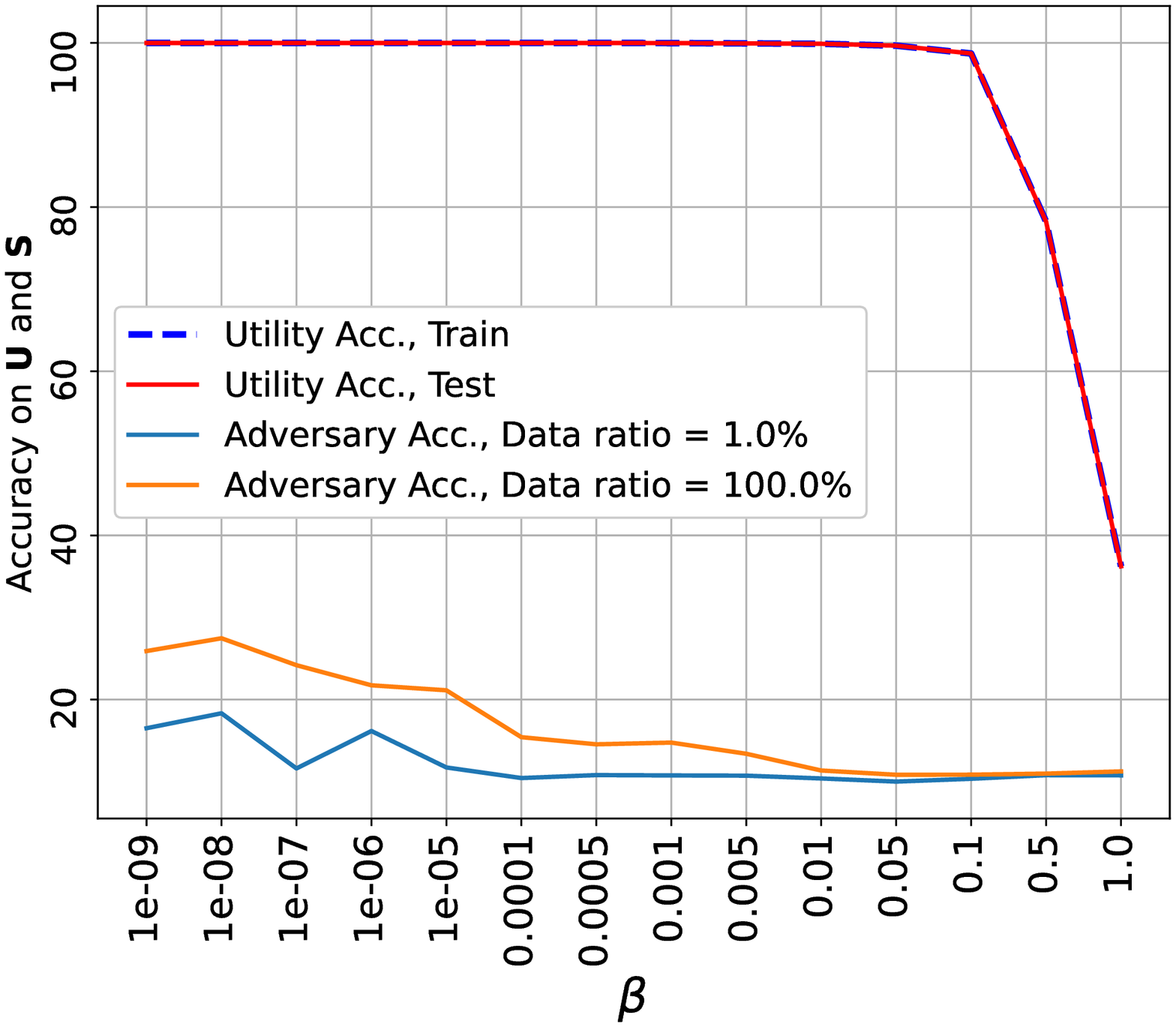}
}
\vfill{}

\subfloat[\label{Fig:ColoredMNIST_g}]{%
  \includegraphics[width=0.34\textwidth,width=0.3\textwidth]{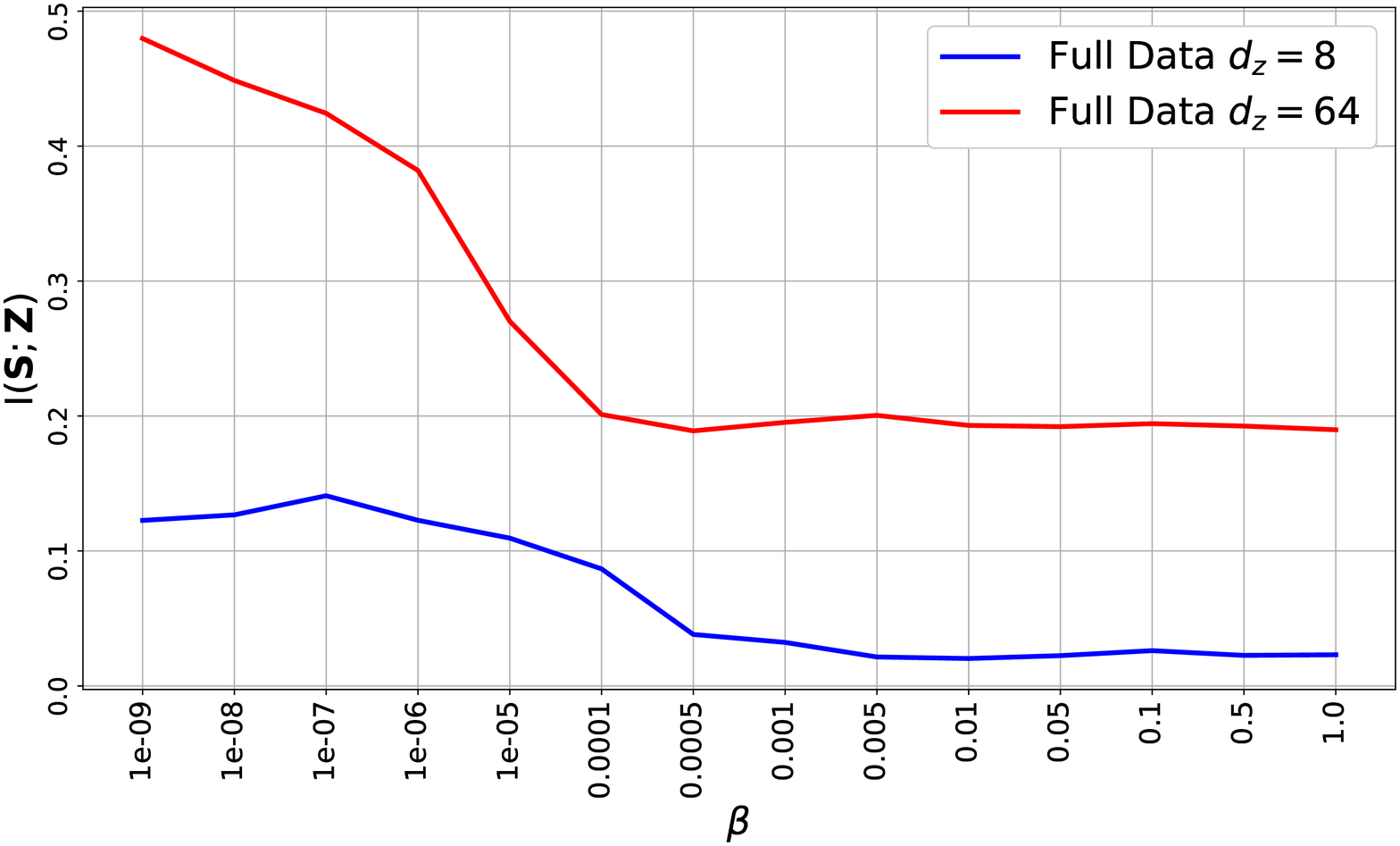}
}
\subfloat[\label{Fig:ColoredMNIST_h}]{%
  \includegraphics[width=0.30\textwidth]{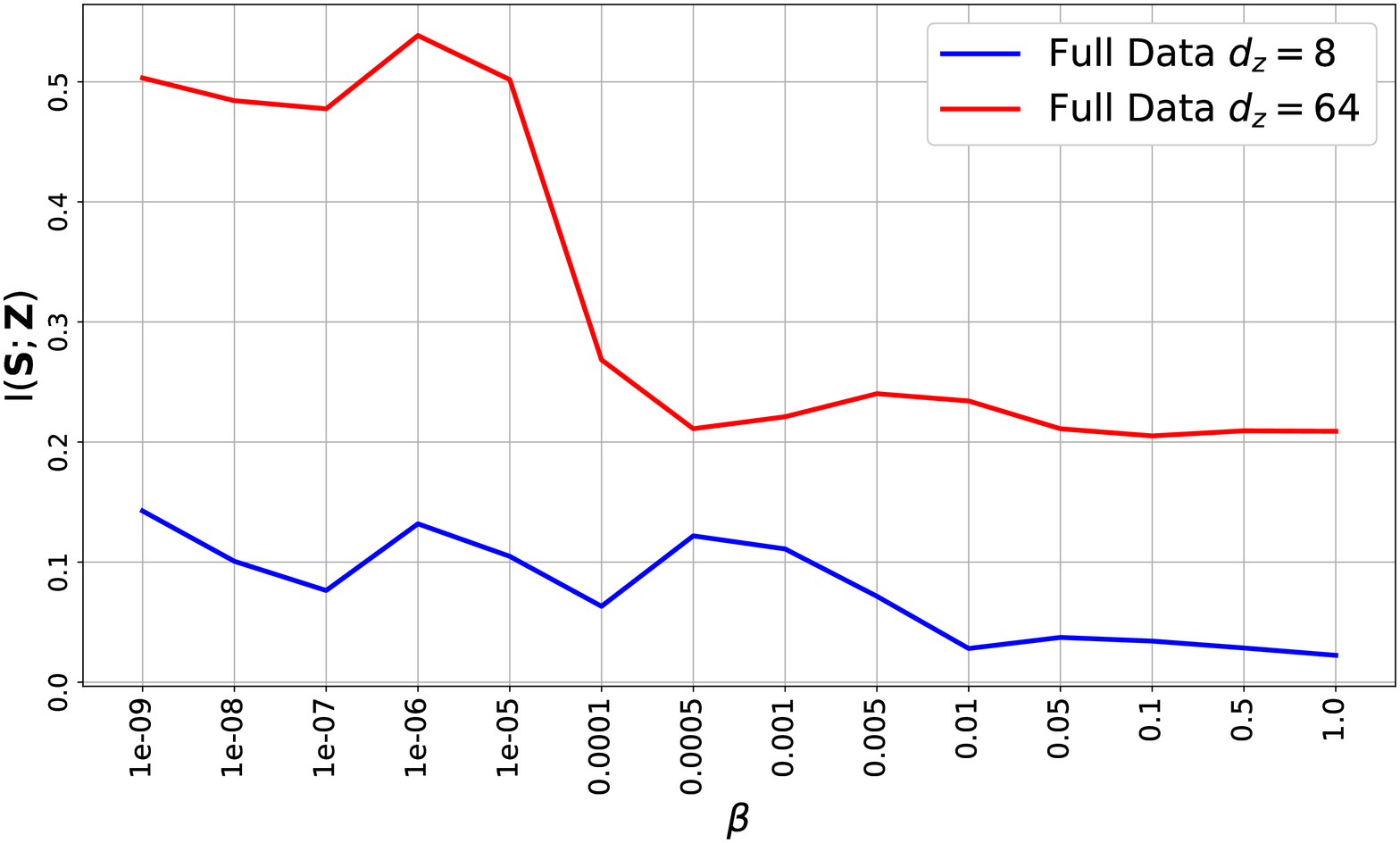}
}
\subfloat[\label{Fig:ColoredMNIST_i}]{%
  \includegraphics[width=0.30\textwidth]{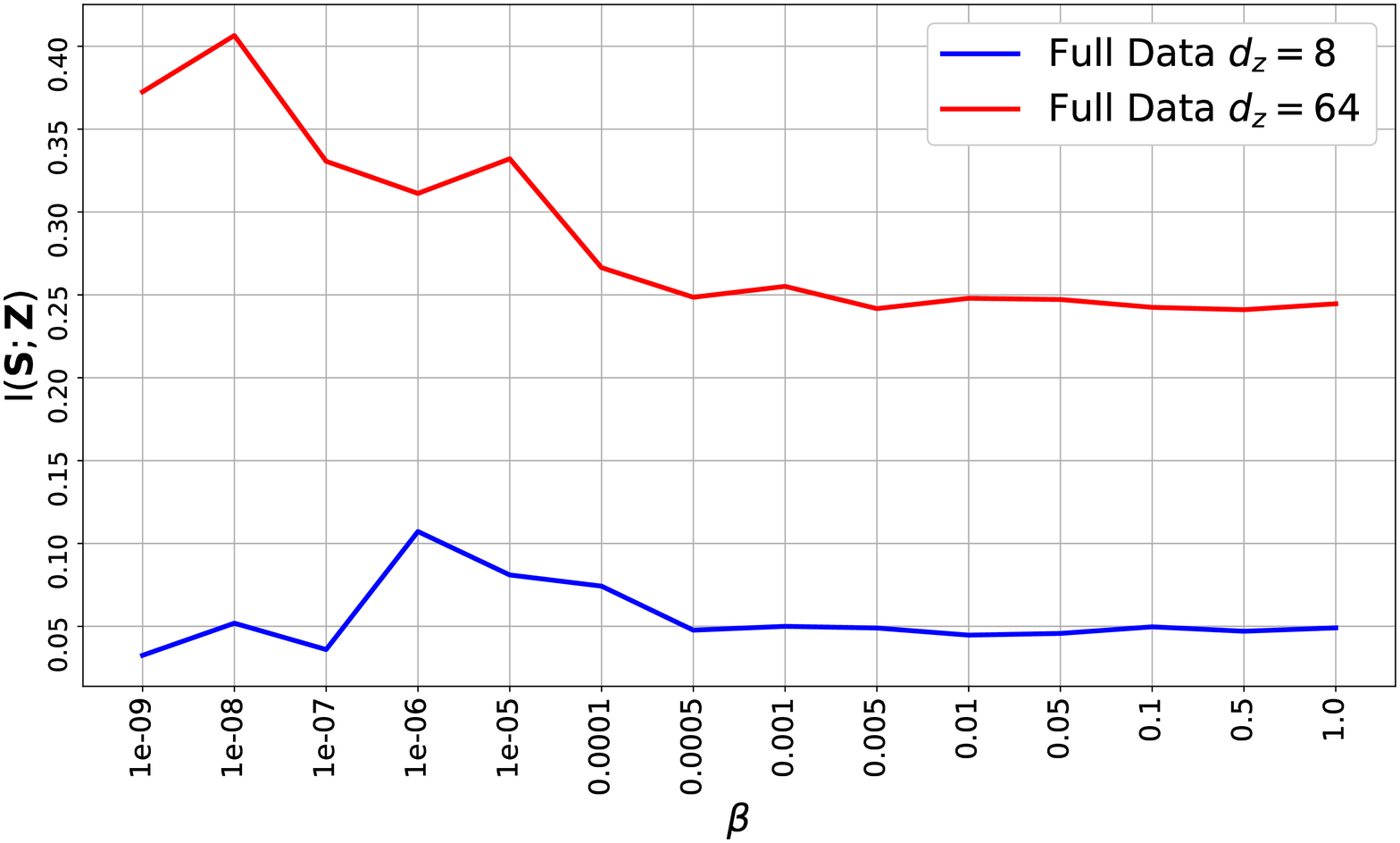}
}
\vfill{}

\subfloat[\label{Fig:ColoredMNIST_j}]{%
  \includegraphics[width=0.30\textwidth]{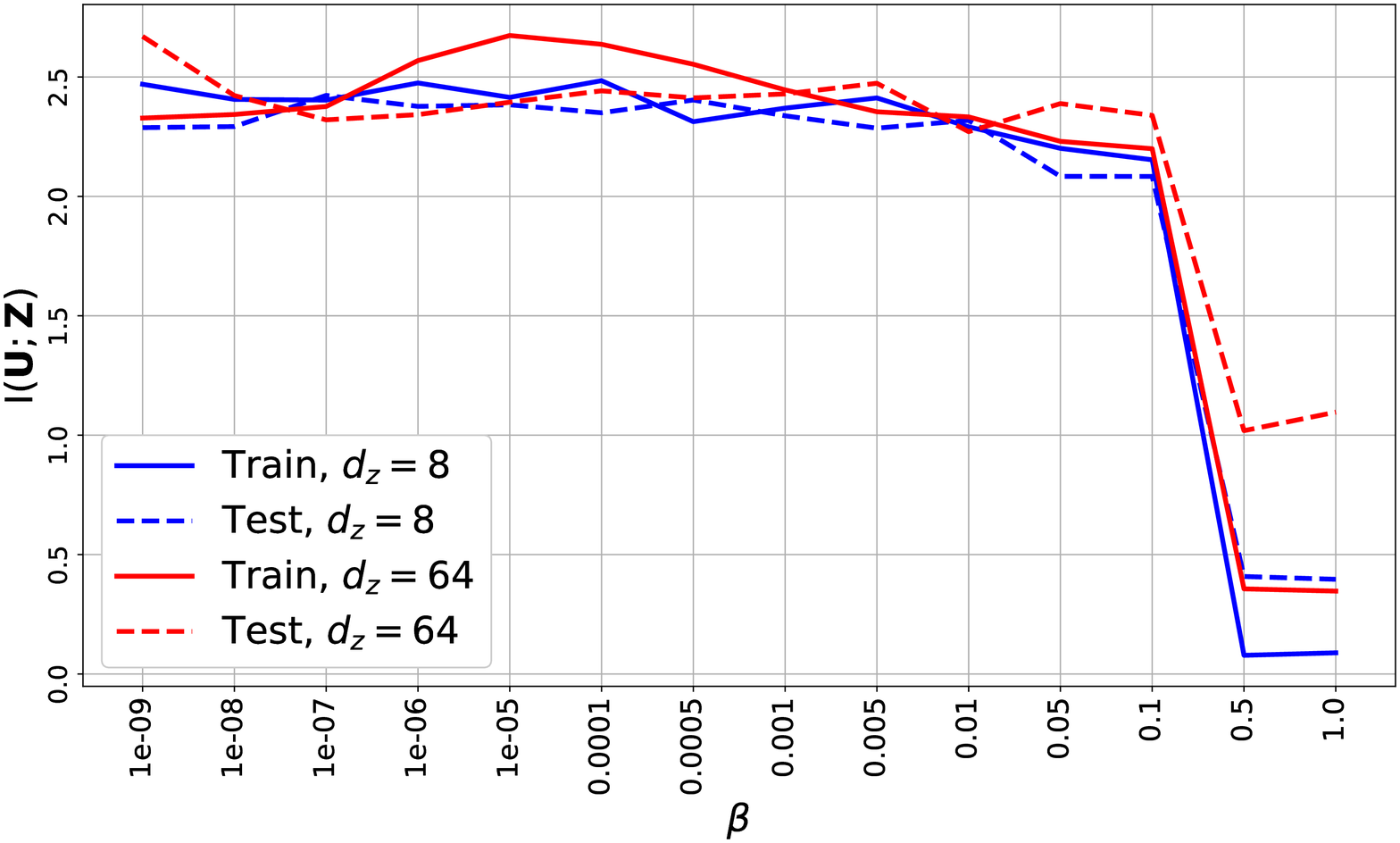}
}
\subfloat[\label{Fig:ColoredMNIST_k}]{%
  \includegraphics[width=0.30\textwidth]{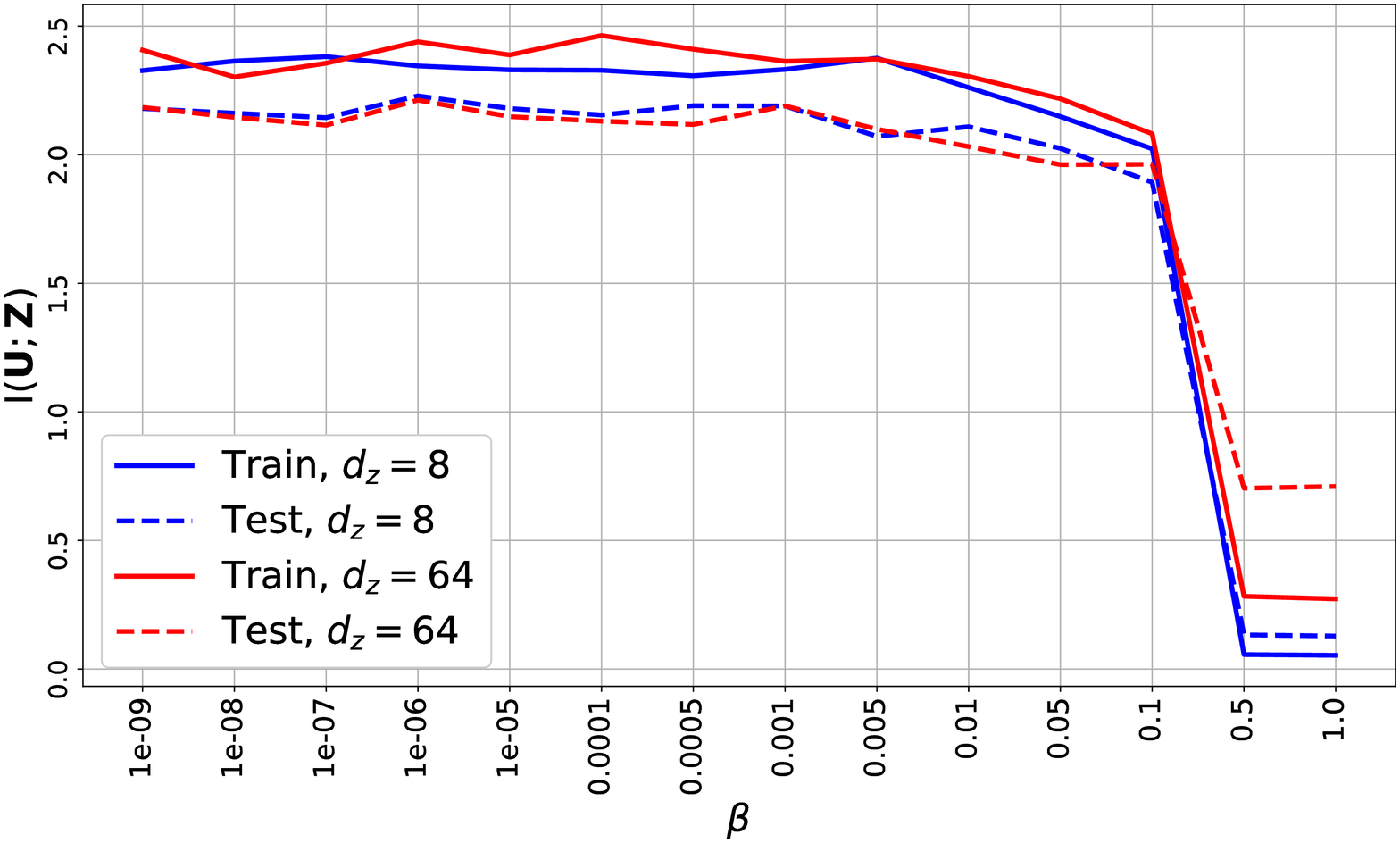}
}
\subfloat[\label{Fig:ColoredMNIST_l}]{%
  \includegraphics[width=0.30\textwidth]{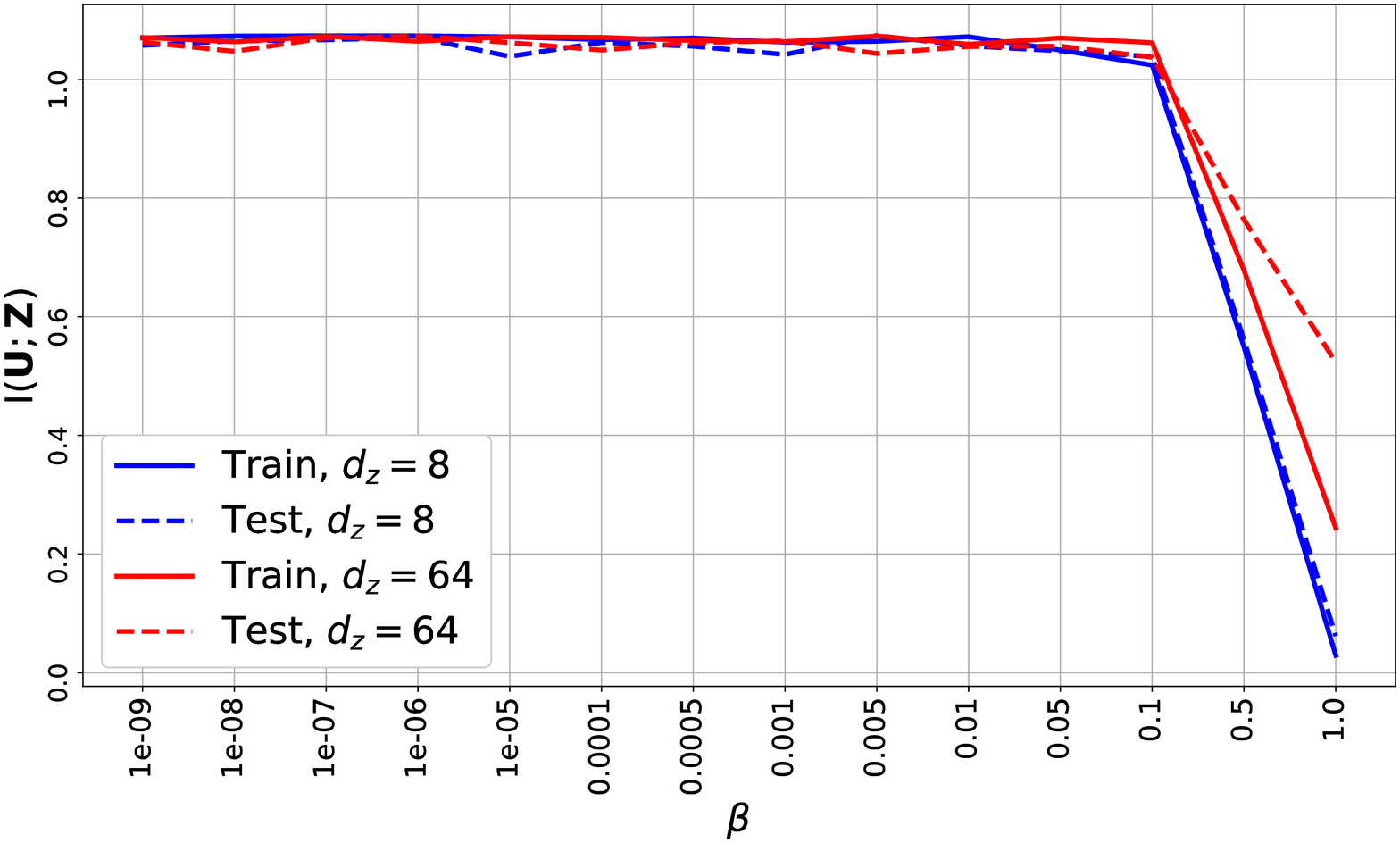}
}
\caption{The results on Colored-MNIST dataset, considering isotropic Gaussian prior. (First Row): $d_{\mathrm{z}} \!= \!8$; (Second Row): $d_{\mathrm{z}} \! = \! 64$; (Third Row): estimated information leakage $\I(\mathbf{S}; \mathbf{Z})$ using MINE; (Fourth Row): estimated useful information $\I(\mathbf{U}; \mathbf{Z})$ using MINE. 
    (First Column): utility task is digit recognition $(\vert \mathcal{U} \vert \!= \! 10)$, while the adversary's goal is the digit color $(\vert \mathcal{S} \vert \! = \! 3)$, setting $P_S (\mathsf{Red}) \!= \!P_S (\mathsf{Green}) \!= \! P_S (\mathsf{Blue}) \! = \! \frac{1}{3}$; 
    (Second Column): utility task is digit recognition $(\vert \mathcal{U} \vert \!= \!10)$, while the adversary's goal is the digit color, setting $P_S (\mathsf{Red}) \! = \! \frac{1}{2}$, $P_S (\mathsf{Green})\! = \! \frac{1}{6}$, $P_S (\mathsf{Blue}) \!=\! \frac{1}{3}$; 
    (Third Column): utility task is digit color recognition $(\vert \mathcal{U} \vert \!=\! 3)$, while the adversary's interest is the digit number $(\vert \mathcal{S} \vert \! = \! 10)$.}\label{Fig:Results_MNIST}
\end{figure*}

In this section, we show the impact of the following factors on the amount of leakage: (i) information complexity regularizer weight $\beta \! \in \! (0, 1 ]$, (ii) released representation dimension $d_{\mathrm{z}}$, (iii) cardinalities of the known utility and unknown sensitive attribute sets, (iv) correlation between the utility and sensitive attributes, and (v) potential bias in a sensitive attribute of adversary's interest.
We conduct experiments on the Colored-MNIST and large-scale CelebA datasets.
The Colored-MNIST\footnote{Several papers have employed Colored-MNIST dataset; However, they are not unique, and researchers synthesized different versions based on their application. The innovative concept behind our version was influenced from the one used in \cite{rodriguez2020variational}.} is \textit{our modified} version of MNIST \cite{lecun-mnisthandwrittendigit-2010}, which is a collection of $70,000$ \textit{colored} digits of size $28  \times  28$. The digits are randomly colored with red, green, or blue based on two distributions, as explained in the caption of Fig.~\ref{Fig:Results_MNIST}. 
The CelebA \cite{liu2015faceattributes} dataset contains $202,599$ images of size $218 \times  178$. 
We used TensorFlow $2.4.1$ \cite{abadi2016tensorflow} with Integrated Keras API. The method details and network architectures are provided in Appendix.~\ref{Appendix:AlgorithmDetails} and Appendix~\ref{Appendix:NetworksArchitecture}.


\begin{figure}[!t]
\centering
\advance\leftskip-0.8cm
\advance\rightskip-0.8cm
\subfloat[\label{Fig:CelebA_a}]{%
  \includegraphics[width=0.26\textwidth]{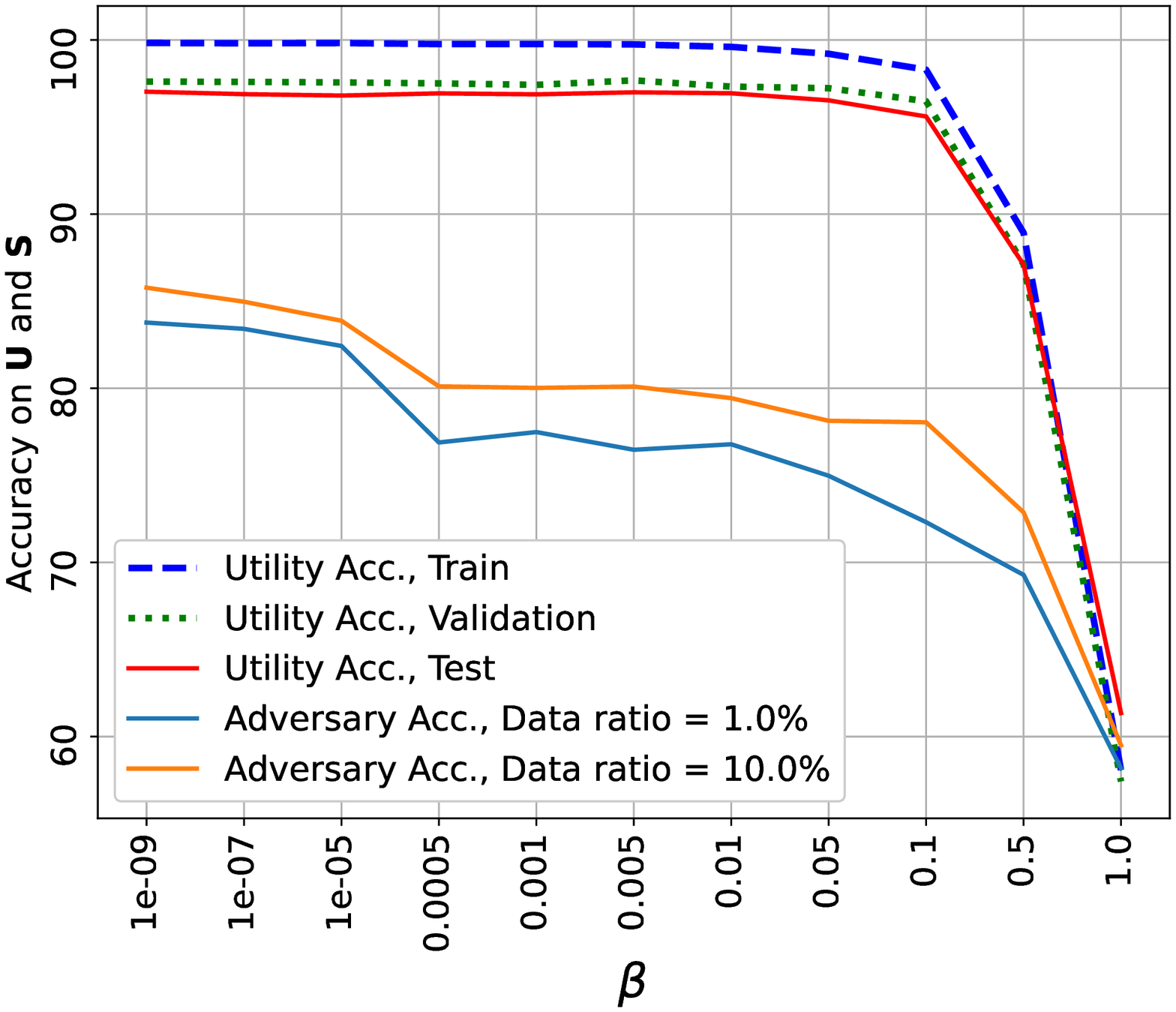}%
}
\subfloat[\label{Fig:CelebA_b}]{%
  \includegraphics[width=0.26\textwidth]{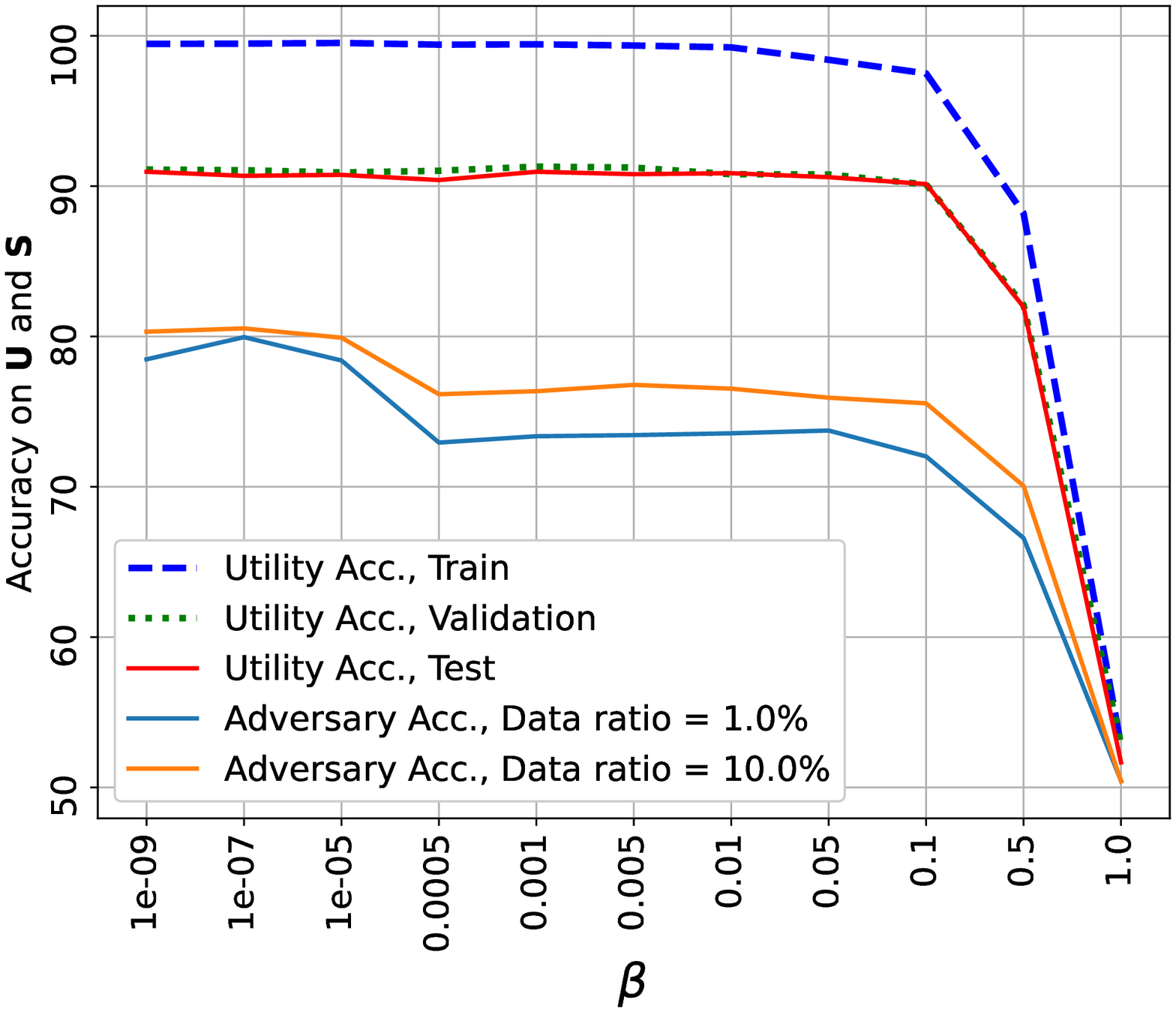}%
}
\vfill{}\vspace{1em}

\subfloat[\label{Fig:CelebA_c}]{%
  \includegraphics[width=0.26\textwidth]{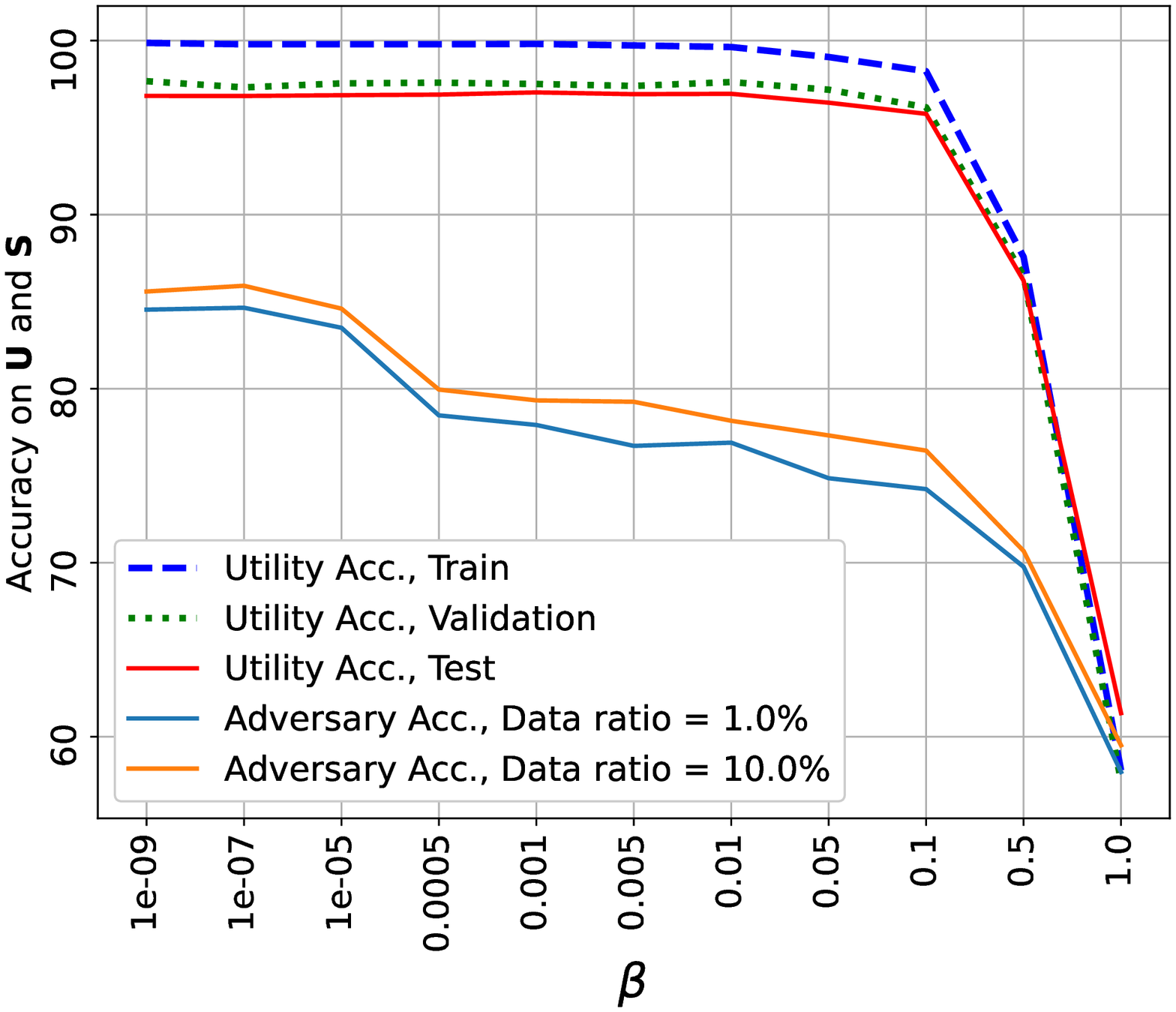}%
}
\subfloat[\label{Fig:CelebA_d}]{%
  \includegraphics[width=0.26\textwidth]{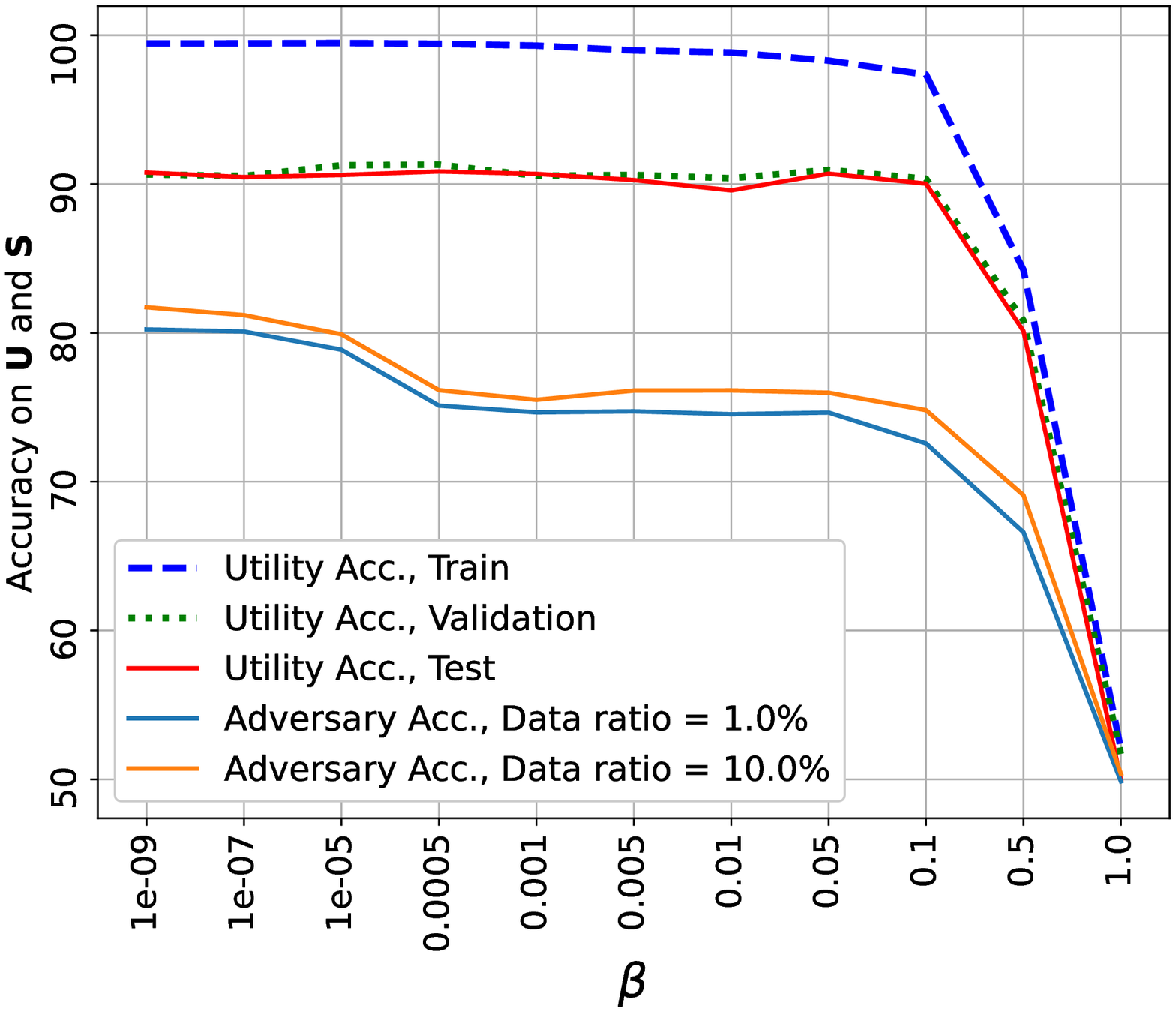}%
}
\vfill{}\vspace{1em}

\subfloat[\label{Fig:CelebA_e}]{%
  \includegraphics[width=0.26\textwidth]{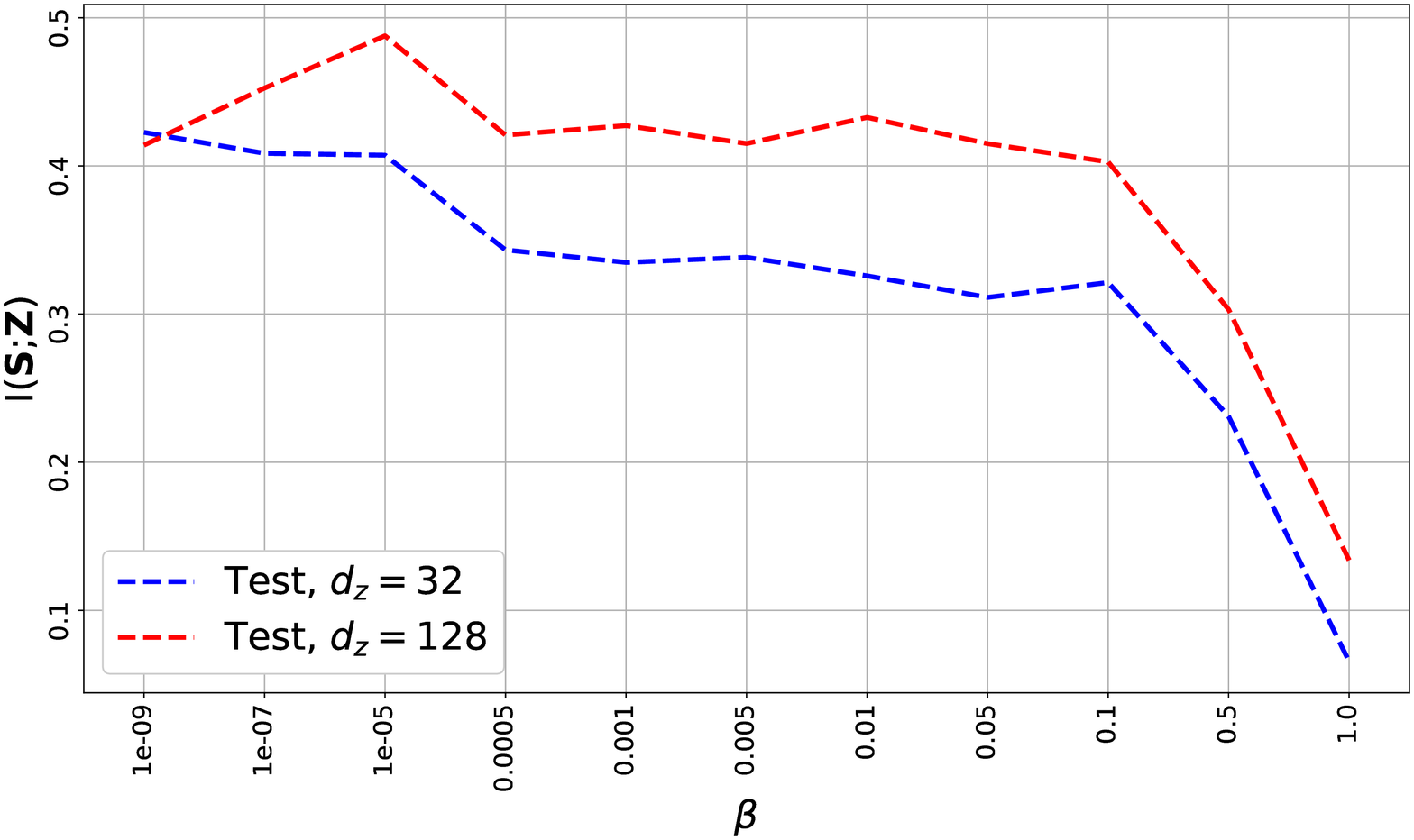}%
}
\subfloat[\label{Fig:CelebA_f}]{%
  \includegraphics[width=0.26\textwidth]{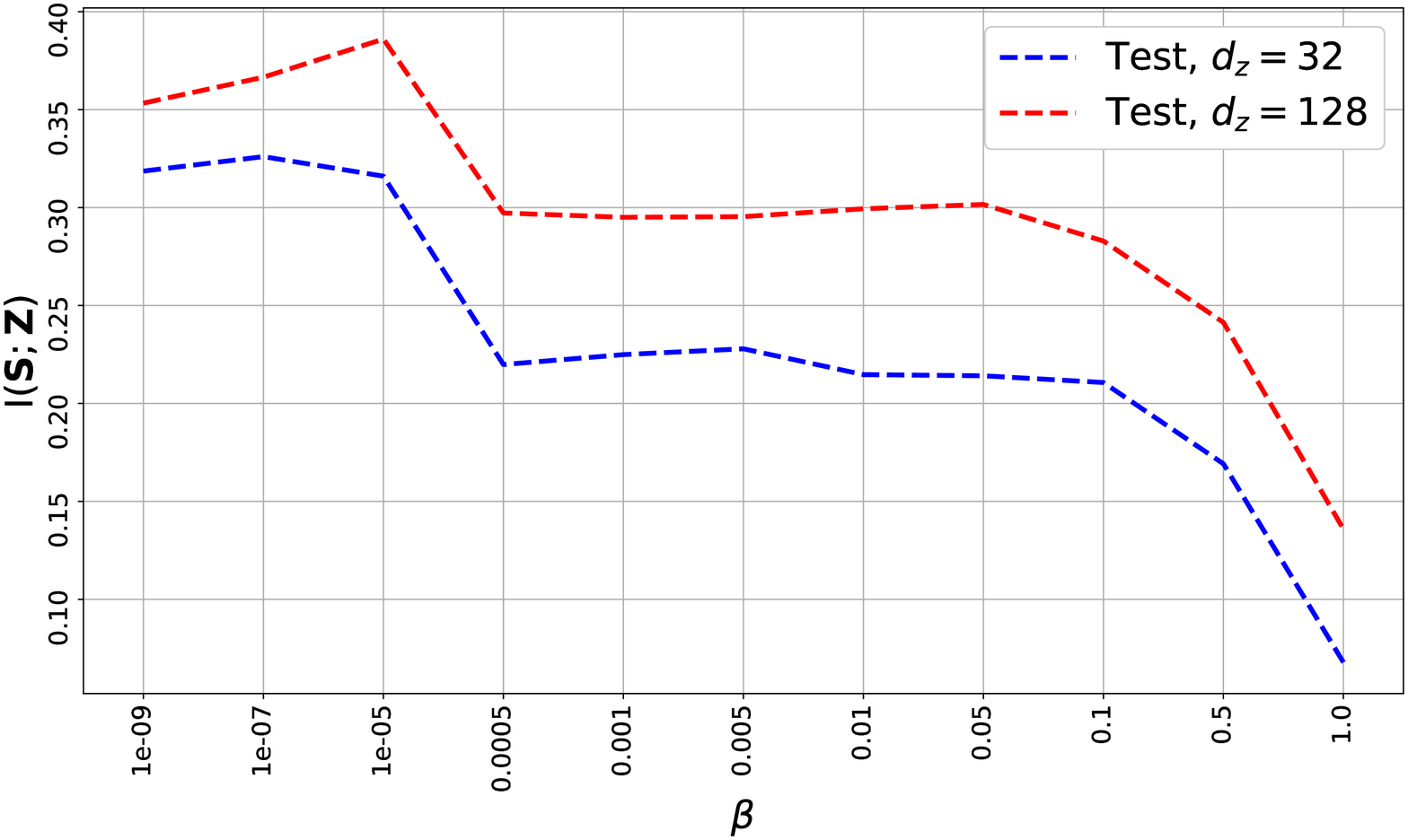}%
}
\vfill{}\vspace{1em}
\subfloat[\label{Fig:CelebA_g}]{%
  \includegraphics[width=0.26\textwidth]{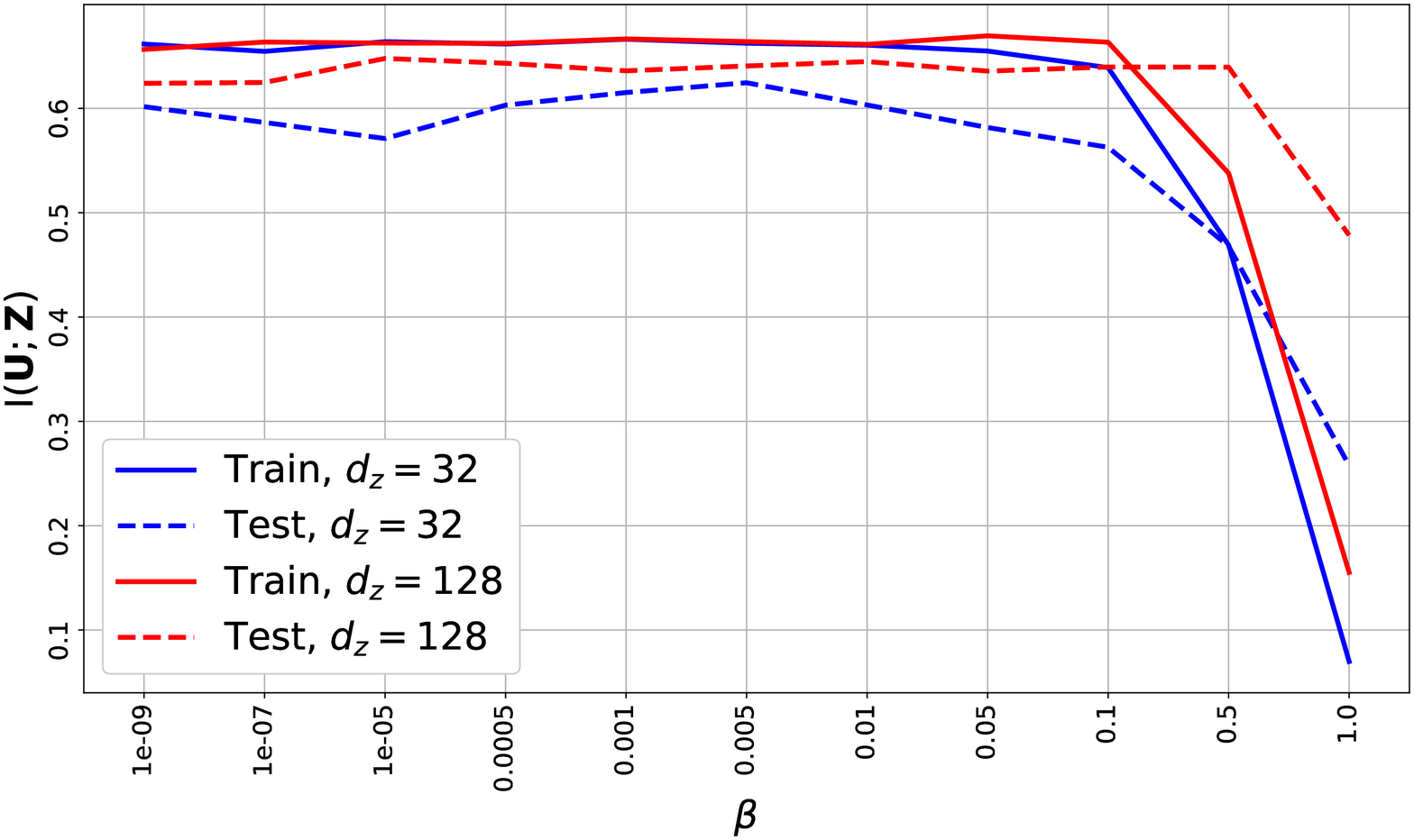}%
}
\subfloat[\label{Fig:CelebA_h}]{%
  \includegraphics[width=0.26\textwidth]{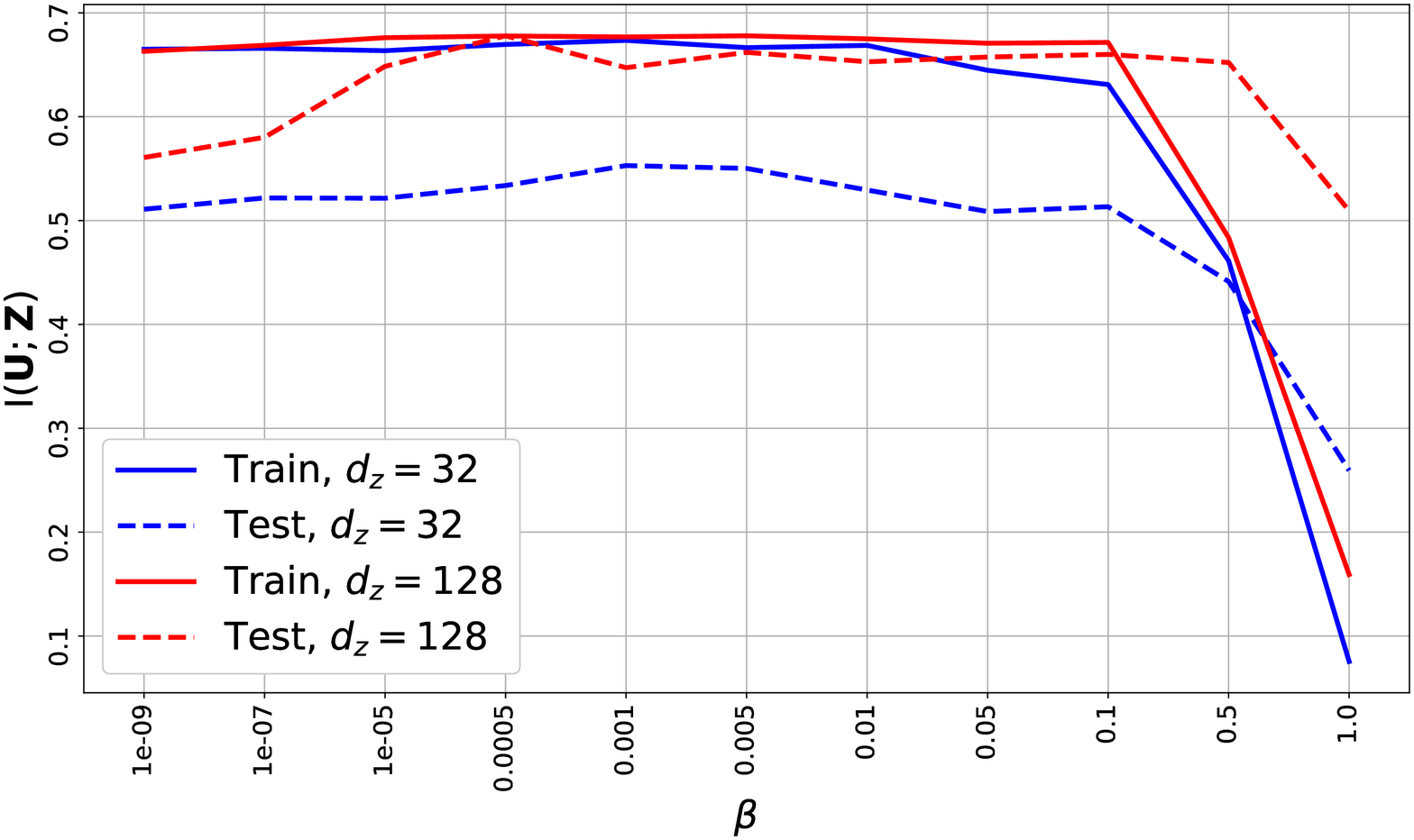}%
}
\vspace{1em}
\caption{The results on CelebA dataset, considering isotropic Gaussian prior. (First Row): $d_{\mathrm{z}} \!= \! 64$; (Second Row): $d_{\mathrm{z}} \!= \! 128$; (Third Row): Estimated information leakage $\I(\mathbf{S}; \mathbf{Z})$ using MINE; (Fourth Row): Estimated useful information $\I(\mathbf{U}; \mathbf{Z})$ using MINE. 
(First Column): utility task is gender recognition $(\vert \mathcal{U} \vert \! =\! 2)$, adversary's interest is heavy makeup $(\vert \mathcal{S} \vert \! =  \! 2)$; 
(Second Column): utility task is emotion (smiling) recognition $(\vert \mathcal{U} \vert \!=\! 2)$, adversary's interest is mouth slightly open $(\vert \mathcal{S} \vert \! =\! 2)$.}
\label{Fig:Results_CelebA}
\end{figure}


The first and second rows of Fig.~\ref{Fig:Results_CelebA} and Fig.~\ref{Fig:Results_MNIST} depict the trade-off among (i) information complexity, (ii) service provider's accuracy on utility attribute $\mathbf{U}$, and (iii) adversary's accuracy on attribute $\mathbf{S}$. 
The third row depicts the amount of information revealed about $\mathbf{S}$, i.e., $\I (\mathbf{S}; \mathbf{Z})$, for the scenarios considered in the first and second rows, which are estimated using MINE \cite{belghazi2018mutual}. 
The fourth row depicts the amount of released information about the utility attribute $\mathbf{U}$, i.e., $\I (\mathbf{U}; \mathbf{Z})$, corresponding to the considered scenarios in the first and second rows, also estimated using MINE. 
We consider different portions of the datasets available for training adversary's network, denoted by the `data ratio'. 
%

The experiments on CelebA consider the scenarios in which the attributes $\mathbf{U}$ and $\mathbf{S}$ are correlated, while $\vert \mathcal{U} \vert = \vert \mathcal{S} \vert = 2$.  We provide utility accuracy curves for (i) training set, (ii) validation set, and (iii) test set. 
As we have argued, there is a direct relationship between information complexity and intrinsic information leakage. Note that, as $\beta$ increases, the information complexity is reduced, and we observe that this also results in a reduction in the information leakage. We also see that the leakage is further reduced when the dimension of the released representation $\mathbf{Z}$, i.e., $d_{\mathrm{z}}$, is reduced. This forces the data owner to obtain a more succinct representation of the utility variable, removing any extra information.

In the Colored-MNIST experiments, provided that the model eliminates all the redundant information $\I (\mathbf{X}; \mathbf{Z} \! \mid \! \mathbf{U})$ and leaves only the information about $\mathbf{U}$, we expect the adversary's performance to be close to `random guessing' since the digit color is independent of its value. We investigate the impact of the cardinality of sets $\vert \mathcal{U} \vert$ and $\vert \mathcal{S} \vert$, as well as possible biases in the distribution of $\mathbf{S}$. The results show that it is possible to reach the same level of accuracy on the utility attribute $\mathbf{U}$, while reducing the intrinsic leakage by increasing the regularizer weight $\beta$, or equivalently, by reducing the information complexity $\I_{\boldsymbol{\phi}} (\mathbf{X}; \mathbf{Z})$. An interesting possible scenario is to consider correlated attributes $\mathbf{U}$ and $\mathbf{S}$ with different cardinality sets $\mathcal{U}$ and $\mathcal{S}$. For instance, utility task $\mathbf{U}$ is personal identification, while the adversary's interest $\mathbf{S}$ is gender recognition.


%
%
%
\section{Conclusion}\label{Sec:Conclusions}

We studied the \textit{variational leakage} to address the amount of potential privacy leakage in a supervised representation learning setup. 
In contrast to the PF and generative adversarial privacy models, we consider the setup in which the adversary's interest is not known a priori to the data owner. We study the role of information complexity in information leakage about an attribute of an adversary interest. 
This was addressed by approximating the information quantities using DNNs and experimentally evaluating the model on large-scale image databases. The proposed notion of \textit{variational leakage} relates the amount of leakage to the minimal sufficient statistics.

\clearpage

\bibliographystyle{ACM-Reference-Format}

\bibliography{references}

\clearpage

%
%

\appendix{\centerline{\textbf{Appendices}}}

\section{Training Details}\label{Appendix:AlgorithmDetails}

All the experiments in the paper have been carried out with the following structure:

\subsubsection{Pre-Training Phase}
\hfill{}\break
We utilize this phase to warm-up our model before running the main training Algorithm~\ref{Algorithm:VariationalNestedLeakage} for the Variational Leakage framework within all experiments. In the warm-up phase, we pre-trained encoder $\left(f_{\boldsymbol{\phi}}\right)$ and utility-decoder $\left(g_{\boldsymbol{\theta}}\right)$ together for the few epochs via \textbf{backpropagation (BP)} with the Adam optimizer \cite{kingma2014adam}. We found out the warm-up stage was helpful for faster convergence. Therefore, we initialize the encoder and the utility-decoder weights with the obtained values rather than random or zero initialization. For each experiment, the hyper-parameters of the learning algorithm in this phase were:

\begin{center}
\resizebox{\linewidth}{!}{%
 \begin{tabular}{||c c c c||} 
 \hline
 Experiment Dataset & Learning Rate & Max Iteration & Batch Size \\ 
 \hline\hline
 Colored-MNIST & \multirow{2}{*}{0.005} & \multirow{2}{*}{50} & \multirow{2}{*}{1024} \\
(both version) &  & &\\ 
 \hline
 CelebA & 0.0005 & 100 & 512 \\
 \hline
\end{tabular}
}
\end{center}

\subsubsection{Main Block-wise Training Phase} \hfill \break
In contrast to the most DNNs training algorithms, each iteration only has one forward step through the network's weights and then update weights via BP approach. Our training strategy is block-wise and consists of multiple blocks in the main algorithm loop. At each block, forward and backward steps have been done through the specific path in our model, and then corresponding parameters update based on the block's output loss path.

%
%
Since it was not possible for us to use the Keras API's default model training function, we implement Algorithm \ref{Algorithm:VariationalNestedLeakage} from scratch in the Tensorflow. It is important to remember that we initialize all parameters to zero except for the $\left( \boldsymbol{\phi}, \boldsymbol{\theta} \right)$ values which acquired in the previous stage. Furthermore, we set the learning rate of the block (1) in the Algorithm \ref{Algorithm:VariationalNestedLeakage}, five times larger than other blocks. The hyper-parameters of the Algorithm \ref{Algorithm:VariationalNestedLeakage} for each experiment shown in the following table:
\begin{center}
\resizebox{\linewidth}{!}{%
 \begin{tabular}{||c c c c||} 
 \hline
 Experiment Dataset & Learning Rate & Max Iteration & Batch Size \\ 
                    &  [blocks (2)-(5)]&           & \\
 \hline\hline 
 Colored-MNIST & \multirow{2}{*}{0.0001} & \multirow{2}{*}{500} & \multirow{2}{*}{2048} \\
 (both version) & & & \\ 
 \hline
 CelebA & 0.00001 & 500 & 1024 \\
 \hline
\end{tabular}
}
\end{center}

\section{Network Architectures}\label{Appendix:NetworksArchitecture}

\subsubsection{MI Estimation}
\hfill{} \break
For all experiments in this paper, we report estimation of MI between the released representation and sensitive attribute, i.e., $\I \left(\mathbf{S};\mathbf{Z}\right)$, as well as the MI between the released representation and utility attribute, i.e., $\I \left(\mathbf{U};\mathbf{Z}\right)$. To estimate MI, we employed the MINE model \cite{belghazi2018mutual}. The architecture of the model is depicted in Table.~\ref{table:Architecture_MINE}.

%
%
%
\begin{table}[t]
\scshape
\small 
\centering
{\renewcommand{\arraystretch}{1.3}%
\begin{tabular}{ p{0.8\linewidth}}
\toprule 
\multicolumn{1}{c}{MINE~~~~$\I \left(\mathbf{U};\mathbf{Z} \right)$} \\
\hline
Input~~$\mathbf{z} \in \mathbb{R}^{d_z}$ Code; $\mathbf{u} \in \mathbb{R}^{\vert \mathcal{U} \vert}$ \\
x = Concatenate([z, u]) \\
FC(100), ELU   \\
FC(100), ELU   \\
FC(100), ELU   \\
FC(1)  \\
\toprule 
\end{tabular}}
\vspace{-3pt}
\caption{The architecture of the MINE network.}
\label{table:Architecture_MINE}
\end{table}

\subsubsection{Colored-MNIST}\hfill \break
In the Colored-MNIST experiment, we had two versions for data utility and privacy leakage evaluation. 
In the first version, we set the utility data to the class' label of the input image and consider the color of the input image as sensitive data, and for the second one, we did vice versa. It is worth mentioning that both balanced and unbalanced Colored-MNIST datasets are applied with the same architecture given in Table~\ref{table:Architecture_Colored-MNIST}.

%
%
%
\begin{table}[h!]
\scshape
\small 
\centering
{\renewcommand{\arraystretch}{1.3}%
\begin{tabular}{ p{0.8\linewidth}}
\toprule 
\multicolumn{1}{c}{Encoder $f_{\boldsymbol{\phi}}$} \\
\hline
Input~~$\mathbf{x} \in \mathbb{R}^{28 \times 28 \times 3}$ Color Image \\
Conv(64,5,2), BN, LeakyReLU   \\
Conv(128,5,2), BN, LeakyReLU   \\
   Flatten  \\
FC($d_z\times4$), BN, Tanh \\
   $\mu$: FC($d_z$), $\sigma$: FC($d_z$)   \\
   $z$=SamplingWithReparameterizationTrick[$\mu$,$\sigma$] \\
\toprule 
\multicolumn{1}{c}{Utility Decoder $g_{\boldsymbol{\theta}}$} \\
\hline
Input~~$\mathbf{z} \in \mathbb{R}^{d_z}$ Code\\
FC($d_z \times 4$), BN, LeakyReLU \\
FC($\vert \mathcal{U} \vert$), SOFTMAX \\
\toprule
\multicolumn{1}{c}{Latent Space Discriminator $D_{\boldsymbol{\eta}}$} \\
\hline
Input~~$\mathbf{z} \in \mathbb{R}^{d_z}$ Code \\
FC($512$), BN, LeakyReLU \\
FC($256$), BN, LeakyReLU \\
FC($1$), Sigmoid \\
\toprule 
\multicolumn{1}{c}{Utility Attribute Class Discriminator $D_{\boldsymbol{\omega}}$} \\
\hline
Input~~$\mathbf{u} \in \mathbb{R}^{\vert \mathcal{U} \vert}$ \\
FC($\vert \mathcal{U} \vert \times 8$), BN, LeakyReLU \\
FC($\vert \mathcal{U} \vert \times 8$), BN, LeakyReLU \\
FC($1$), Sigmoid \\
\toprule 
\end{tabular}}
\vspace{-3pt}
\caption{The architecture of DNNs used in for the Colored-MNIST experiments.}
\label{table:Architecture_Colored-MNIST}
\end{table}

%
\subsubsection{CelebA}
\hfill{} \break
In this experiment, we considered three scenarios for data utility and privacy leakage evaluation, as shown in Table~\ref{table:CelebA_Scenarios}. 
Note that all of the utility and sensitive attributes are binary. 
The architecture of the networks are presented in Table~\ref{table:Architecture_CelebA}.

\begin{table}[t!]
\resizebox{\linewidth}{!}{%
 \begin{tabular}{||c c c||} 
 \hline
Scenario Number & Utility Attribute & Sensitive Attribute\\
 \hline\hline 
 1 & Gender & Heavy Makeup\\
 \hline
 2 & Mouth Slightly Open & Smiling\\
 \hline
 3 & Gender & Blond Hair\\
 \hline
\end{tabular}
}
\caption{Scenarios considered for CelebA experiments.}
\vspace{-3pt}
\label{table:CelebA_Scenarios}
\end{table}
%
%

\vfill{}

%
%
%
\begin{table}[t!]
\scshape
\small 
\centering
{\renewcommand{\arraystretch}{1.3}%
\begin{tabular}{ p{0.82\linewidth}}
\toprule 
\multicolumn{1}{c}{Encoder $f_{\boldsymbol{\phi}}$} \\
\hline
Input~~$\mathbf{x} \in \mathbb{R}^{64 \times 64 \times 3}$ Color Image \\
Conv(16,3,2), BN, LeakyReLU   \\
Conv(32,3,2), BN, LeakyReLU   \\
Conv(64,3,2), BN, LeakyReLU   \\
Conv(128,3,2), BN, LeakyReLU   \\
Conv(256,3,2), BN, LeakyReLU   \\
Flatten  \\
FC($d_z\times4$), BN, Tanh \\
$\mu$: FC($d_z$), $\sigma$: FC($d_z$)   \\
$z$=SamplingWithReparameterizationTrick[$\mu$,$\sigma$] \\
\toprule 
\multicolumn{1}{c}{Utility Decoder $g_{\boldsymbol{\theta}}$} \\
\hline
Input~~$\mathbf{z} \in \mathbb{R}^{d_z}$ Code\\
FC($d_z$), BN, LeakyReLU \\
FC($\vert \mathcal{U} \vert$), SOFTMAX \\
%
\toprule 
\multicolumn{1}{c}{Latent Space Discriminator $D_{\boldsymbol{\eta}}$} \\
\hline
Input~~$\mathbf{z} \in \mathbb{R}^{d_z}$ Code \\
FC($512$), BN, LeakyReLU \\
FC($256$), BN, LeakyReLU \\
FC($1$), Sigmoid \\
\toprule 
\multicolumn{1}{c}{Utility Attribute Class Discriminator $D_{\boldsymbol{\omega}}$} \\
\hline
Input~~$\mathbf{u} \in \mathbb{R}^{\vert \mathcal{U} \vert}$ \\
FC($\vert \mathcal{U} \vert \times 4$), BN, LeakyReLU \\
FC($\vert \mathcal{U} \vert$), BN, LeakyReLU \\
FC($1$), Sigmoid \\
\toprule 
\end{tabular}}
\vspace{-3pt}
\caption{The architecture of networks for the CelebA experiments.}
\label{table:Architecture_CelebA}
\end{table}
\begin{figure}[!ht]
\centering
\advance\leftskip-5cm
\advance\rightskip-5cm

\subfloat[Autoencoder module\label{Fig:TF_Autoencoder}]{%
  \includegraphics[width=0.5\textwidth,height=0.2\textheight]{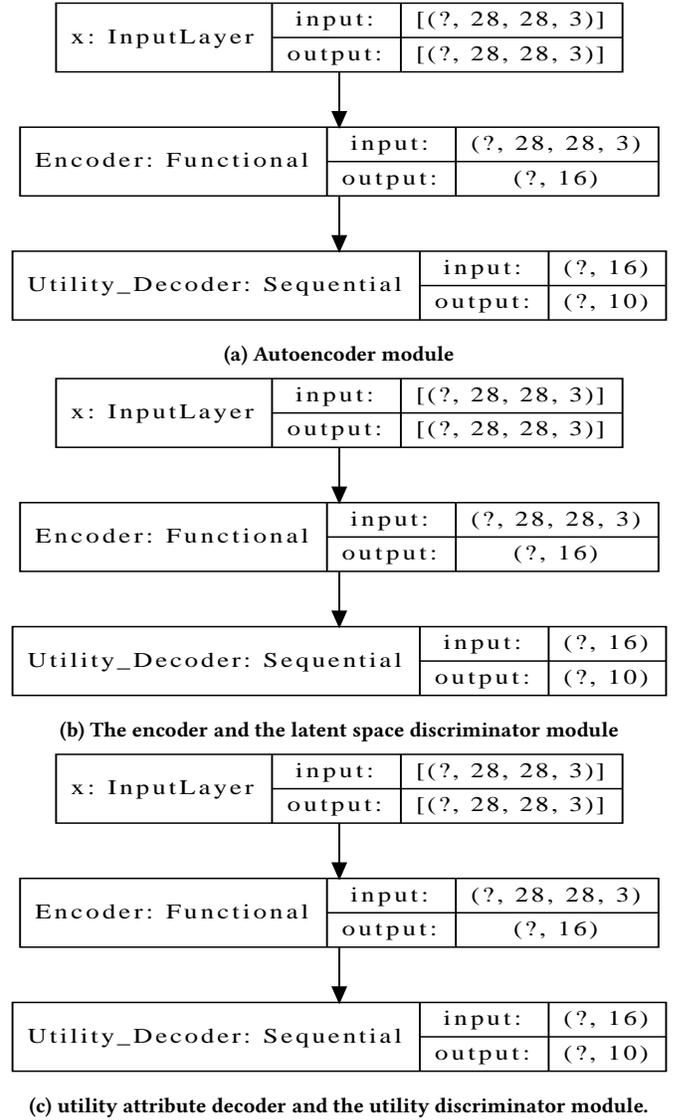}
}
\vfill{}

\subfloat[The encoder and the latent space discriminator module\label{Fig:TF-encoder_zdiscriminator}]{%
  \includegraphics[width=0.5\textwidth,height=0.2\textheight]{image/TF-ae}
}
\vfill{}

\subfloat[utility attribute decoder and the utility discriminator module.\label{Fig:TF-encoder_zdiscriminator}]{%
  \includegraphics[width=0.5\textwidth,height=0.2\textheight]{image/TF-ae}
}
\caption{The three sub-modules that make up the main network.}\label{Fig:TF_sub_modules}
\end{figure}

\section{Implementation Overview}\label{Appendix:ImplementationOverview}

\begin{figure*}[!htb]
\centering
\advance\leftskip-0.8cm
\advance\rightskip-0.8cm
\includegraphics[width=1.06\textwidth]{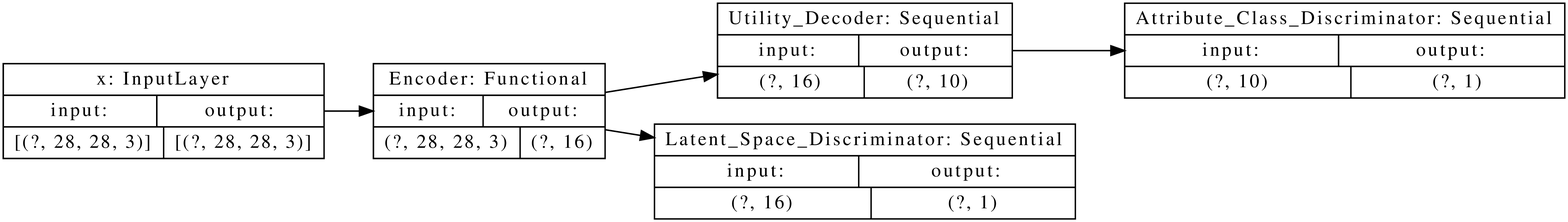}
\caption{Complete model structure in the training phase. The above model defines for Colored-MNIST dataset with the number of classes as utility attributes and digit colors as sensitive attributes. Also, the encoder output is set to 16 neurons. (The adversary model is not part of the data owner training algorithm)}
\label{Fig:TF_full_model}
\end{figure*}

Fig.~\ref{Fig:TF_full_model} demonstrates all sub networks of the proposed framework that attached together and their parameters learned in the Algorithm \ref{Algorithm:VariationalNestedLeakage}.

However, we did not define and save our model in this form because of technical reasons to efficiently implement the training algorithm. As shown in Algorithm~\ref{Algorithm:VariationalNestedLeakage}, the main loop consists of five blocks where only some networks are used in the forward phase, and mostly one of them would update their parameters via BP in each block. Therefore, we shattered the model into three sub-modules in the training stage for simplicity and performance. Fig.~\ref{Fig:TF_sub_modules} shows the corresponding sub-modules of Fig.~\ref{Fig:TF_full_model}, which are used in our implementation. During training, all of the sub-module (a) parameters, call "autoencoder part" would update with BP after each forward step. For the (b) and (c) sub-modules, only the parameters of one network are updated when the corresponding error function values backpropagate, and we freeze the other networks parameters in the sub-module. For example, block (3) of Algorithm~\ref{Algorithm:VariationalNestedLeakage} is related to the (b) sub-module, but at the BP step, the latent space discriminator is frozen to prevent its parameters from updating. This procedure is vice versa for module (b) at block (2).

It should be mentioned that during our experiments, we found out that before running our main algorithm, it is beneficial to pre-train the autoencoder sub-module since we need to sample from the latent space, which uses in other parts of the main model during training. We justify this by mentioning that sampling meaningful data rather than random ones from latent variables from the beginning of learning helps the model to converge better and faster in comparison with starting Algorithm \ref{Algorithm:VariationalNestedLeakage} with a randomly initiated autoencoder.

\end{document}